\documentclass[runningheads]{llncs}

\usepackage[mobile]{eccv}

\usepackage{eccvabbrv}
\usepackage{algorithm}
\usepackage{algpseudocode}
\usepackage{graphicx}
\usepackage{booktabs}
\usepackage{amsmath}
\usepackage{threeparttable}
\usepackage{adjustbox}
\usepackage{fvextra}

\usepackage[accsupp]{axessibility}  

\usepackage[pagebackref,breaklinks,colorlinks,citecolor=eccvblue]{hyperref}
\usepackage{hyperref}

\usepackage{orcidlink}

\usepackage{multirow}
\usepackage{float}

\begin{document}


\title{Robusto-2: Benchmarking Humans \& VLMs for Autonomous Driving in Lima \& New York City}


\titlerunning{Robusto-2}

\author{Adrian Cespedes \and
Marcelo Chincha, \\ 
Dunant Cusipuma \and
Victor Flores-Benites \and
David Ortega \and
Arturo Deza
}

\authorrunning{Cespedes, Chincha et al.}

\institute{Artificio \\Lima, Peru \\
\email{\{adrian.cespedes,marcelo.chincha, dunant.c,vfloresb,david.ortega,deza\}@artificio.org}\\
\url{https://artificio.org}}

\maketitle

\begin{abstract}
As autonomous vehicles (AVs) continue to expand internationally and use multi-modal systems such as VLMs as a cognitive backbone for their Action models; how well will these systems generalize in new settings, in particular out-of-distribution (OOD) edge-case scenarios in new geographies? In this paper, we study this open question by providing a full factorial analysis with participants from Lima, participants from New York City, and VLMs and showing them dashcam footage collected from Lima and New York City -- prompting them with a variety of questions under a Visual Question Answering (VQA) paradigm. In particular, we pick these two cities as they are highly challenging driving locations with no commercial AV deployment, and ask questions that span 4 categories: \textit{Factual}, \textit{Ratings}, \textit{Counterfactual} and \textit{Reasoning}. We find that Humans and VLMs diverge in their responses -- though this is modulated by the type of questions asked, and that Humans answer similarly independent of where they are from (Lima/NYC). To our surprise, we did not find a strong difference in terms of answers (Humans or VLMs) that was modulated by geography, likely due to their high out-of-distribution nature. Our dataset is available at ~\href{https://huggingface.co/datasets/Artificio/robusto-2}{https://huggingface.co/datasets/Artificio/robusto-2}
\end{abstract}

\section{Introduction}
\label{sec:intro}

How well do VLMs cross generalize to new driving scenarios? Today, the algorithms powering AV technology have slowly been evolving ranging from hard-coded heuristics~\cite{moravec2005stanford,thorpe1988vision,cui2021lookout}, to end-to-end models~\cite{chen2024end,chib2023recent} and now reversing to a middle-ground of end-to-end and yet modular systems such as Vision-Language-Action (VLA) models powered by Vision-Language models (VLMs) as cognitive backbones~\cite{sima2024drivelm,renz2025simlingo,zhou2026autovla}. Despite this exciting and fast-paced evolution in academia, very little research has been focusing on evaluating how such models perform within the real world, and more specifically to different geographies and chaotic environments~\cite{cusipuma2025robusto,zhou2024vision,gao2026steervla,xie2025vlms}.


In this paper we are motivated to answer the previous question, but also within the context of how markets evolve in the deployment and adaptation of AV technology. As of today, New York City (USA) and Lima (Peru), rank among the most challenging cities to drive in within their respective regions~\cite{McCann_2025, Cantu_2024}, and are both cities without a commercial AV deployment -- limited testing with safety drivers notwithstanding -- likely both due to their complexity and regulation (unlike San Francisco, London, Dubai or Shenzhen where several startups have already started deploying their fleets). In that regard, it is important to perform studies assessing the cross-cultural generalization capacity of these systems because most multi-modal foundation models~\cite{nazi2025vision,liu2025culturevlm,cusipuma2025robusto}  -- be it VLMs or VLAs -- are trained with Western Hemisphere data, mainly from the United States~\cite{mehrabi2021survey,sun2025uncovering}. This raises an important \textit{safety} question: when these systems are later deployed in other parts of the world~\cite{huang2025vlms} -- in particular to locations where little to no data is collected for training or validation, how well will they generalize? And how big is this cross-cultural generalization gap, if it exists?

\begin{figure}[t]
    \centering
    \includegraphics[width=\textwidth]{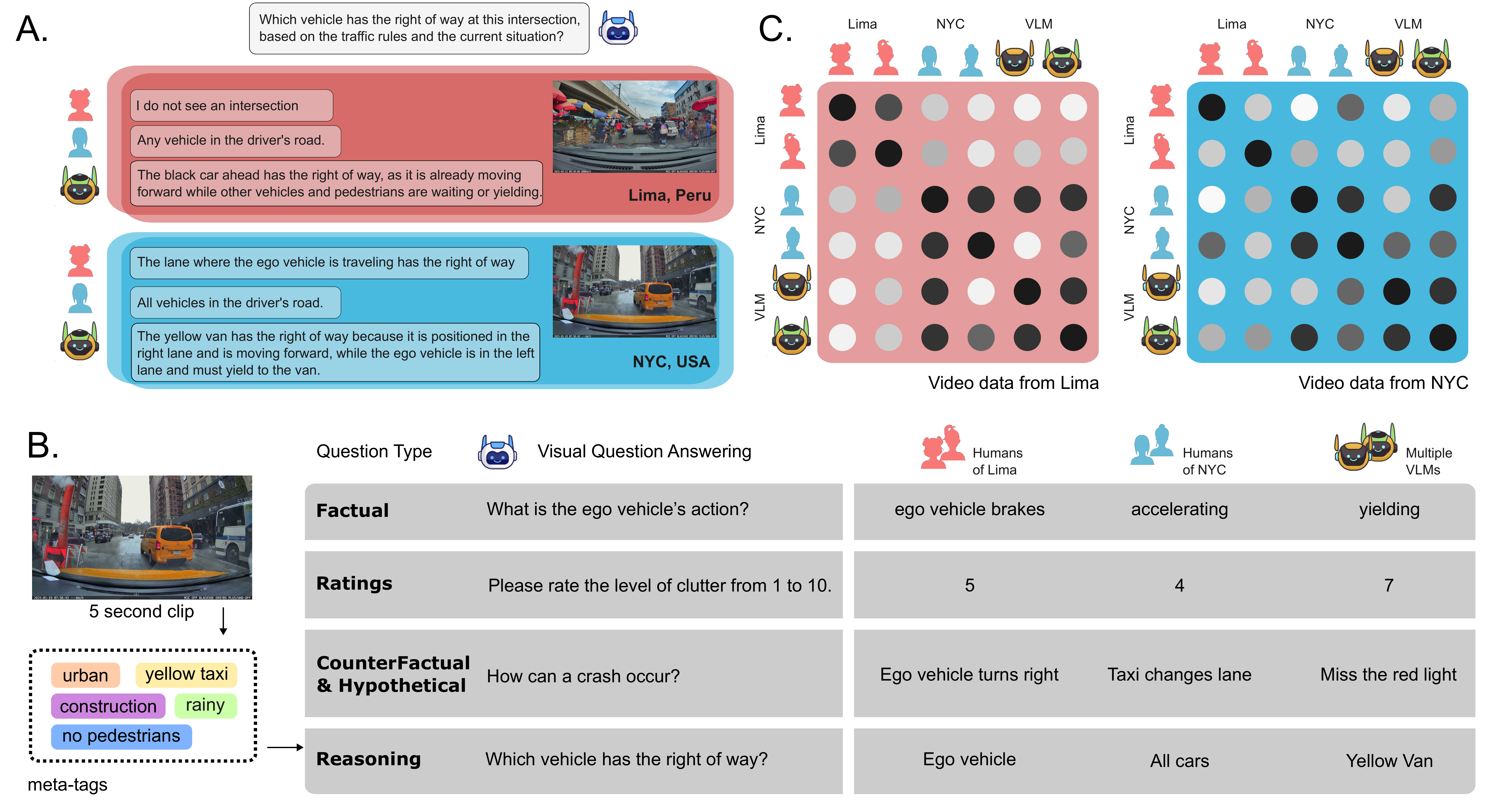}
    \caption{How well do Humans from Lima, Humans from NYC and VLMs compare to each other cognitively when being shown Out-of-Distribution video clips (\textit{A.})? In this paper we study the cross-cultural (Lima vs NYC) and cross-system (Human vs VLM) generalization gap when prompted with data from both geographies. Studying this gap under a VQA framework (\textit{B.}) is important for explainability purposes and as AVs continue to expand outside pioneering cities like San Francisco and Shenzhen. In particular we will aim to disentangle interactions between the individuals and video data from both Lima \& NYC \textit{(C.)}.}
    \label{fig:Figure1}
\end{figure}

To answer this question we introduce \textbf{Robusto-2}, our benchmark for studying the cross-cultural (Lima \textit{vs} NYC) and cross-system (Human \textit{vs} VLM) generalization gap in autonomous-driving VQA. We collect a series of dashcam driving videos recorded from the \textit{same camera} from both Lima and New York City and perform Visual Question Answering~\cite{antol2015vqa} on a collection of 3 populations: humans from Lima, humans from New York City and VLMs released by different start-ups \& organizations---though, as we quantify later, not all of them are architecturally independent, as shown in Figure~\ref{fig:Figure1}. This allows us to create a fully factorial design of 2 $\times$ 3 where we can study how answers compare across humans of these 2 cities and VLMs contingent on \textit{where} the data was recorded from, and to \textit{whom} the questions were asked. Our data, stimuli and analysis code are released at \href{https://huggingface.co/datasets/Artificio/robusto-2}{huggingface.co/datasets/Artificio/robusto-2}.


\section{Experimental Setup}
Previous work such as Robusto-1\cite{cusipuma2025robusto} has shown that Humans and VLMs either converged or diverged in their answers contingent on the types of questions asked (Factual, Multiple-Choice or Counter-Factual \& Hypotheticals) when prompted \textit{O.O.D.} video clips from Lima, Peru. Furthermore, there has been more recent work that has benchmarked several VLMs models within the context of Autonomous Driving and also using Visual Question Answering (VQA)~\cite{antol2015vqa}, as it allows us to `cognitively probe' each system given their answer~\cite{marcu2024lingoqa,kuznietsov2024explainable,atakishiyev2024explainable}. In this paper we will extend the analysis by adding a control group from New York City: both participants and video data; as one could perhaps imagine that it is unfair to compare humans from Lima (a non-US city with non-native english speakers) to VLMs trained in US-based data as was done in Robusto-1~\cite{cusipuma2025robusto}. This nuance is important because questions in the previously mentioned study were asked in english. We provide details of our experimental setup in the rest of this section.

\subsection{Participants}

A total of 10 humans from Lima, 10 humans from New York City and 10 VLMs were used (totalling 30 systems). Throughout the paper we use \emph{system} for an individual respondent (one participant or one VLM, hence $30$ systems) and \emph{system type} for the three-level factor of our design (Lima participants / NYC participants / VLMs); \emph{geography} always refers to where the \emph{video} was recorded. All questions were asked in english to all systems for the Visual Question Answering (VQA) procedure.

Human participants were recruited through an open call published on our organization's official LinkedIn account and digital online ads targeting active university students aged between 18 and 30 years. Interested candidates completed a pre-registration screening survey. To proceed, candidates were required to verify their student status and hold a valid driver's license for their respective cities.

Eligible candidates then entered a verification and trial phase. They submitted their official documentation and completed a pilot trial consisting of VQA tasks for three videos from Robusto-1~\cite{cusipuma2025robusto}. This trial served to familiarize participants with the task dynamics and verify the English proficiency of non-native speakers (specifically from Lima). Candidates who failed to meet the verification criteria or did not complete the trial were disqualified, and replacements were selected from the pre-registration pool. In conditions where the candidate pool exceeded the required sample size, a random lottery was utilized for selection.

All data collection was hosted via Google Forms. To bypass global character response limits imposed by the platform, the evaluation was split into two sequential forms containing 10 videos each, maintaining the exact question structure and unique presentation order. Access to the forms was restricted to verified emails provided during pre-registration.

Specific logistical challenges, variable recruitment timelines, detailed compensation structures (including geographic adjustments and referral bonuses), and specific driver's license requirements are further detailed in the Supplement.

\subsection{Clips extraction}

\begin{figure}[!t]
    \centering
    \includegraphics[width=\textwidth]{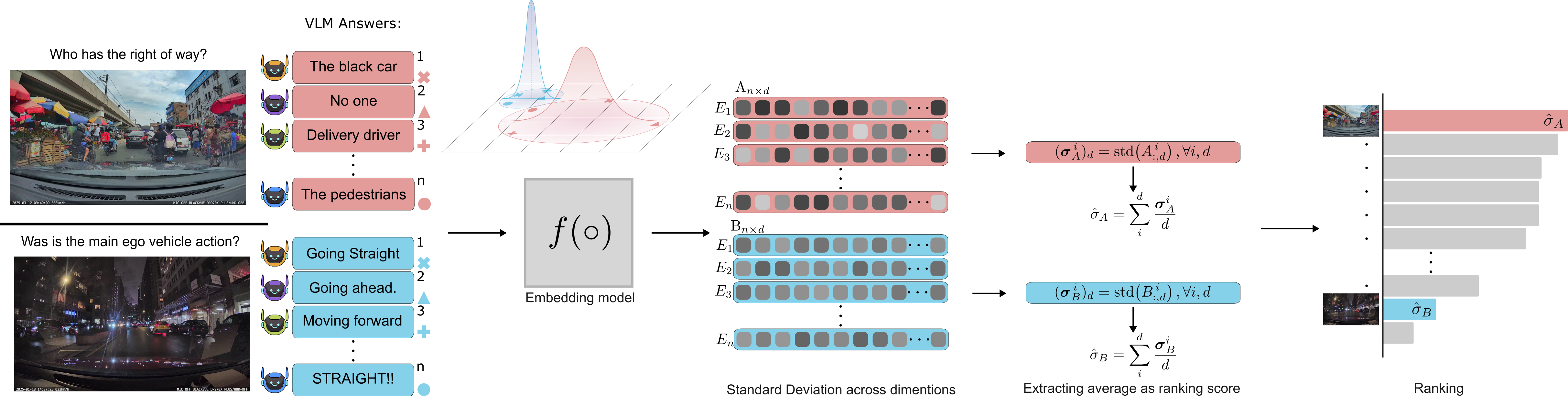}
    \caption{A diagram explaining the OOD selection pipeline: 200 video clips from Lima and NYC are fed to multiple VLMs by asking them the same 20 questions via VQA. We then select the 10 clips of Lima and 10 clips of NYC that have the highest variance in answers across models, interpreting this \textit{disagreement} as OOD scenes. These clips are then shown to Humans in our survey.}
    \label{fig:Figure2}
\end{figure}

The raw video recordings were processed by extracting the most relevant segments, resulting in a total of 200 5-second clips (half recorded in New York City and the other half in Lima). Each clip was then annotated with metadata describing its key characteristics, including the ego-vehicle’s actions, the behavior of surrounding vehicles and pedestrians, the state of traffic lights, as well as sky and weather
conditions. From these annotated clips, the first block of video-specific questions was automatically generated, while the remaining three blocks were predefined.

Based on the VQA responses, embeddings were computed and used to perform a dispersion-based analysis across videos. For every (video, question) pair we measured how far the models' answers spread apart from one another: we took the standard deviation \emph{across models} along each embedding dimension and averaged it over dimensions, then averaged that quantity over the five questions of each block to obtain one score per video and block. Videos were ranked in descending order of this score, so that a high rank means the models disagreed most about that clip. This procedure enabled the identification of out-of-distribution (OOD) samples by highlighting videos with higher dispersion across their embedding space. The final twenty clips were then picked greedily down these rankings under two constraints: an equal split of ten Lima and ten NYC clips, and the exclusion of clips temporally overlapping an already-selected one (many clips originate from the same continuous recording), so that no scene is represented twice. Because blocks are ranked independently and a clip can only be taken once, the four blocks contribute unequal numbers of clips. The resulting OOD videos were then selected to poll humans from NYC and humans from Lima. The full procedure is detailed in Sec.~\ref{sec:supp-ood}. This pipeline is summarized in Figure~\ref{fig:Figure2} and Figure~\ref{fig:Figure1} \textit{(B.)}.

\subsection{Questions for Visual Question Answering (VQA)}
\label{sec:VQA}

In the following sub-section we will provide details of the type of questions asked in the Visual Question Answering experiment for Humans and VLMs. A full list of the 20 questions can be seen in the Supplementary Material.


\underline{\textbf{Block 1: Factual (Q1 -- Q5)}} These questions were factual questions that were related to weather, time of day, traffic signals and flow. For example, for one video; two different samples of 5 questions may look like:
Examples of these questions asked in the survey were: 
%

\textbf{Q1}: \textit{What is the ego vehicle's current action?}

\textbf{Q2}: \textit{Why is the vehicle maintaining a straight path?}




\medskip\noindent
While for another video the first five questions may look like:

\textbf{Q1}: \textit{What action is the ego vehicle taking?}

\textbf{Q2}: \textit{Why is the vehicle braking?}


\medskip\noindent
It is important to note that these questions \textit{varied} per each video and were consistently different. In other words, every video had 5 unique factual questions, but every participant (Human or VLM) saw the same set of questions per video.

\medskip\noindent
\underline{\textbf{Block 2: Ratings (Q6 -- Q10)}} These are questions that we ask across all systems to test how they quantitatively assess several quantities such as clutter, hazard and probabilistic outcomes -- more generally to see how they make perceptual and cognitive judgments. These questions are fixed for all videos, two of the five can be seen below:

\textbf{Q6} : \textit{Please rate the level of clutter from 1 to 10. Consider 10 as the highest level of clutter and 1 as the lowest.}

\textbf{Q7} : \textit{On a scale from 1 (not likely) to 10 (very likely), how likely is it that you encounter a similar driving scenario in your regular driving experience?}



\medskip\noindent
\underline{\textbf{Block 3: CounterFactual \& Hypothetical (Q11 -- Q15)}} These questions are generated to explore the creativity of both Humans and VLMs, as these questions objectively do not have a proper ground truth. Here are 2 of the 5 questions we systematically ask across all systems.

\textbf{Q11} : \textit{What would have had to happen in this video for a crash to have occurred involving the driver?}

\textbf{Q12} : \textit{What would have had to happen in this video for an external crash to have occurred not involving the driver?}




\medskip\noindent
\underline{\textbf{Block 4: Reasoning (Q16 -- Q20)}} This block is an extension that was not included in Robusto-1~\cite{cusipuma2025robusto}, as models now include long Chain-of-Thought primitives. To identify how similar all systems reason (Humans vs VLMs), we ask them these reasoning questions -- once again many of these questions are ill-posed and difficult as it has been suggested that VLMs still struggle~\cite{chen2025bring,kumar2025reinforcing}. Two of these five questions can be read below:

\textbf{Q16} : \textit{Which vehicle has the right of way at this intersection, based on the traffic rules and the current situation?}

\textbf{Q17} : \textit{Is it currently safe for pedestrians to cross the street? Please explain your reasoning.}




\begin{figure}[t]
    \centering
    \includegraphics[width=\textwidth]{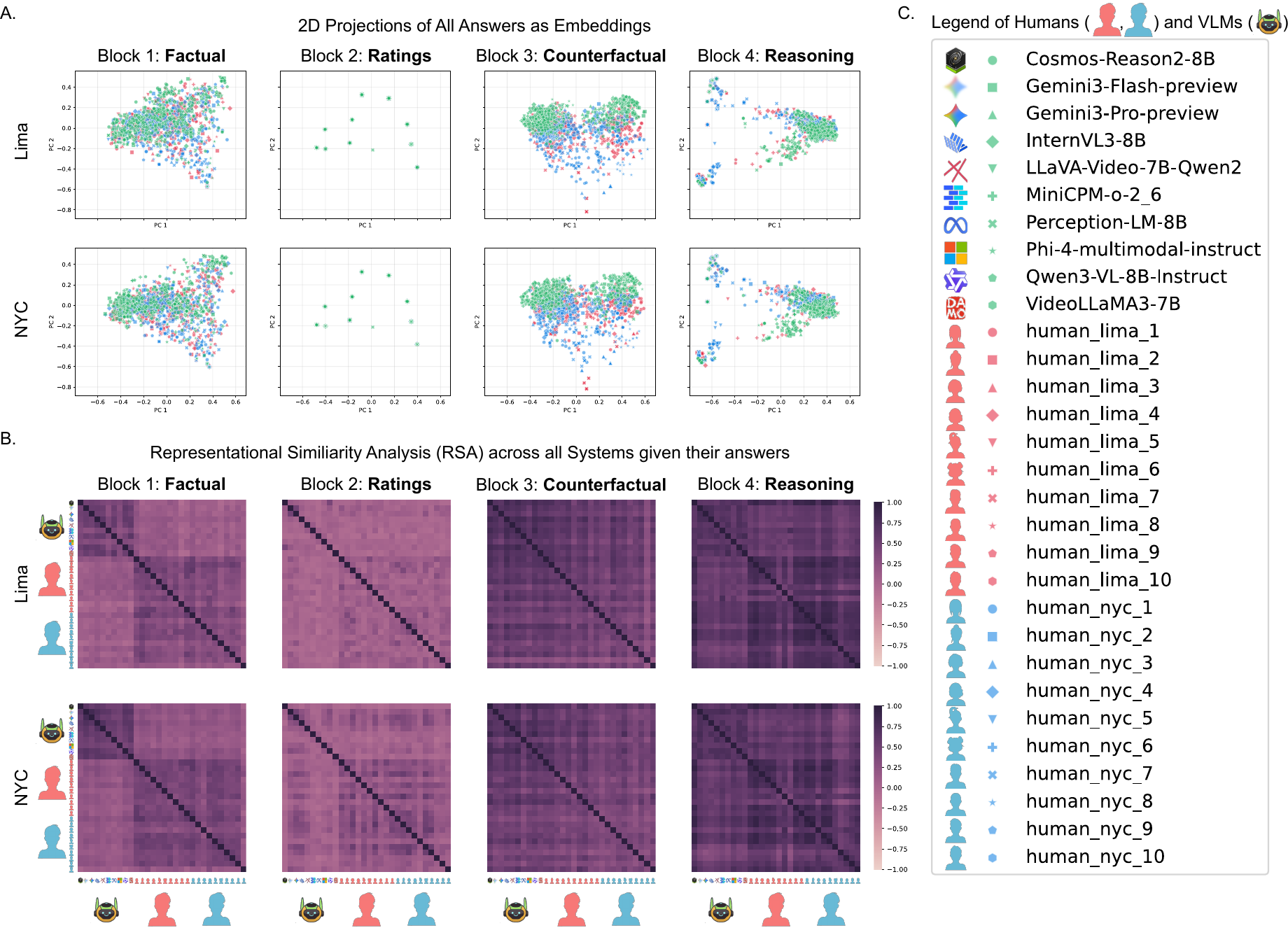}
    \caption{An assessment of how Humans and VLMs converge/diverge depending on the questions asked and the geography of both the data and the participant for Lima and NYC locations. In \textit{(A.)} we see the raw answers projected in 2D via PCA in a block-wise fashion to directly compare Lima vs NYC per block, in \textit{(B.)} we see the Representational Similarity Matrix (RSA) across all systems -- Humans \& VLMs are shown in \textit{(C.)}. }
    \label{fig:Assessment}
\end{figure}

\section{General Assessment of Responses for Humans and Machines across Lima and New York City}
\label{sec:General}

The first analysis we performed revolved around knowing the differences between humans from Lima, humans from NYC, and VLMs. To accomplish this we projected all the answers through sentence embeddings for each system to a 2D plot via PCA and also performed Representational Similarity Analysis~\cite{kriegeskorte2008representational,kornblith2019similarity} on such embeddings to intuitively show the similarities and differences across all systems per block of questions (Figure~\ref{fig:Assessment}).

Generally, we observe little to no difference of pattern of results whether the data is from Lima or NYC, but it is rather modulated by the type of questions asked across Humans and VLMs (\textit{eg.} Block 1 vs Block 2). We believe the differences across geographies are subtle because the questions asked steer the space of answers in a narrower subspace. The PCA plot shown in Figure~\ref{fig:Assessment}
(A.) high-lights these similarities as block-wise plots of the raw embeddings. In addition we also ran a global PCA analysis and these projections can be seen in the Supplementary Material. Performing both of these analysis is relevant to inspect the local block-wise and global differences of the answer even if the projections are linear. We also replicated this analysis with another embedding model (\texttt{Qwen3-Embedding-4B}~\cite{zhang2025qwen3}) instead of \texttt{all-mpnet-base-v2}~\cite{song2020mpnet} and see how the general pattern of results hold as observed in the Supplement.

\section{Analysis of Systematic Bias} 

\begin{figure}[!t]
    \centering
    \includegraphics[width=\textwidth]{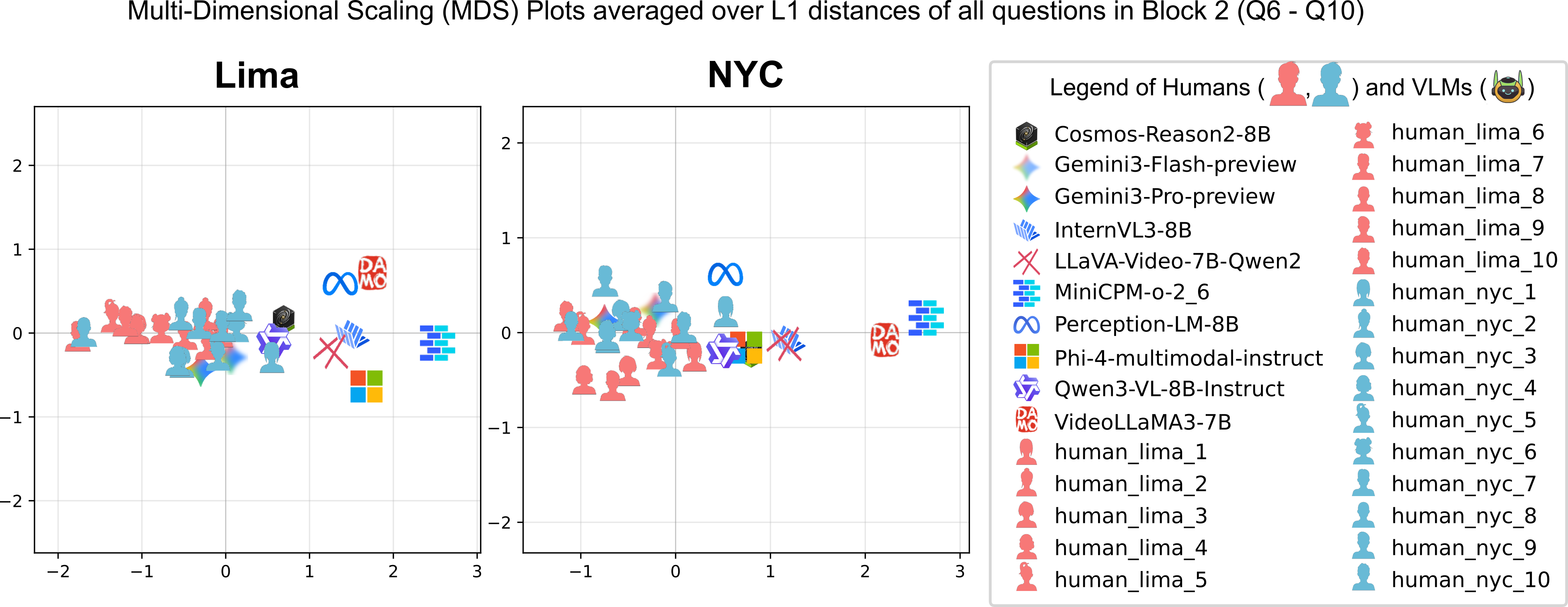}
    \caption{An averaged \textit{per} Geography (Lima vs NYC) Multi-Dimensional Scaling plot computed via the L1 distances of ratings across all questions of Block 2 (Q6 - Q10) showing all Humans and VLMs. Notice that the geometry of ratings is very similar across Geographies, and that Humans are very similar to each other, but not so much to VLMs -- however, this Human vs VLM difference is accentuated in Lima.}
    \label{fig:MDS_total}
\end{figure}

Our next analysis consisted of studying how Humans and VLMs responded to the set of 5 questions we have that involved a 1 to 10 rating scale. Recall from Block 2, that these questions are \textit{subjective} by nature, and were asked across \textit{all videos} to each VLM and Human as follows:

\begin{itemize}
\item \textbf{Q6:} Please rate the level of clutter from 1 to 10. Consider 10 as the highest level of clutter and 1 as the lowest.
\item \textbf{Q7:} On a scale from 1 (not likely) to 10 (very likely), how likely is it that you encounter a similar driving scenario in your regular driving experience?
\item \textbf{Q8:} On a scale from 1 (not hazardous) to 10 (very hazardous), how hazardous is the situation for the driver?
\item \textbf{Q9:} On a scale from 1 to 10, how well do you think an autonomous vehicle would handle driving in this scene? Consider 1 as very poor driving and 10 as perfect driving.
\item \textbf{Q10:} On a scale from 1 to 10, how well do you think you could handle driving in this scene? Consider 1 as very poor driving and 10 as perfect driving.
\end{itemize}


\begin{figure}[!t]
    \centering
    \includegraphics[width=\textwidth]{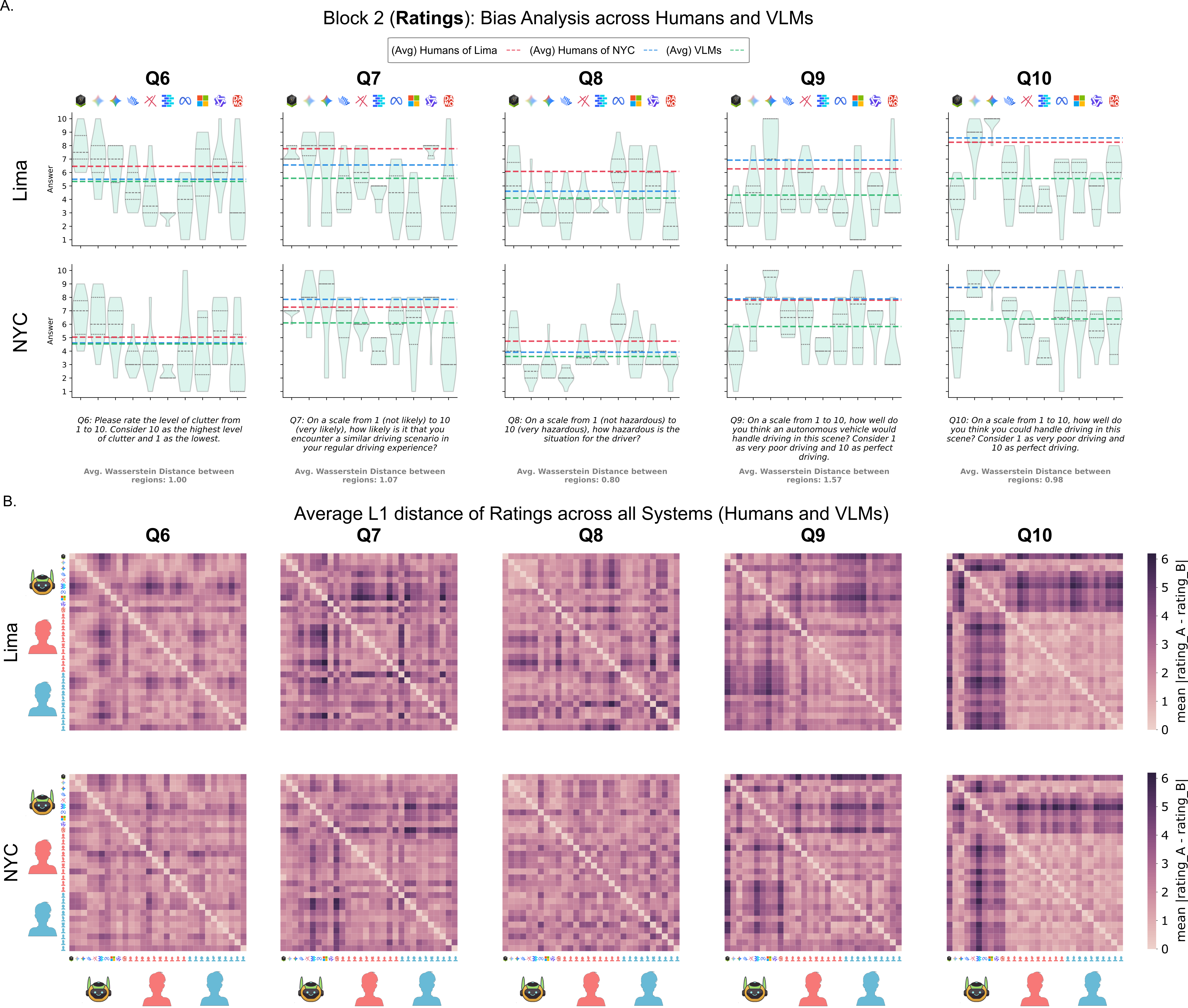}
    \caption{Analysis of Systematic bias covered through distributions \textit{(A.)} and inter-system differences per video \textit{(B)}. In \textit{(A.)} we compute the average ratings of each system via 1 repetition, and compare these to violin plots computed for each VLM over the 20 repetitions per video. In \textit{(B.)} we compute the L1 distance between all pairs of each system's ratings, and compute an average across all videos per question. In both plots, there is high similarity across systems when shown videos from either Lima or NYC.}
    \label{fig:BiasAnalysis}
\end{figure}

The first analysis we did was to perform a Multi-Dimensional Scaling procedure on the \textit{dissimilarity} matrix computed by the raw numeric scores across all systems per video. This difference is computed using the L1 distance of each rating and done in a fully factorial manner across all systems pairs. The \textit{per} Geography average of such results is shown in Figure~\ref{fig:MDS_total} where we observe that Humans cluster together independently of where the data is from (Lima or NYC). We do find that VLMs however seem to be further away from the Human cluster in Lima than in New York City, and that Humans from New York City seem to have less of a difference with VLMs than Humans from Lima in both geographies.

We then wanted to analyze the pattern of ratings as distributions of each system (Humans and VLMs), and in particular for VLMs that when prompted multiple times may provide different ratings. To produce a distribution, we made each VLM respond to every video-question pair 20 times, we used these repeated measurements to construct compound violin plots capturing the distribution of ratings. The responses were first separated by video region (Lima on the top row and NYC on the bottom row). For each of the five questions, we generated a dedicated subplot. In each subplot, every violin represents the full distribution of ratings (including the 20 repetitions) produced by a specific VLM for that question within a given region (e.g., all ratings from Cosmos-Reason2-8B for Question 6 across Lima videos). This result can be seen in Figure~\ref{fig:BiasAnalysis} \textit{(A.)}.
Additionally, each subplot includes two human consensus references, one for Lima (red) and one for NYC (blue), computed as the average rating across participants for each corresponding human region, video region, and question combination. We also added a VLM consensus (green) by averaging over the 1st repetition of the 20 used to produce the distribution across videos to control for the number of points used to compute such distribution.

We then calculated the Wasserstein distance averaged across all violin plots averaged over the VLMs. These are reported in the lower row of the plot. We noticeably found that the most similar pattern of ratings across VLMs was for Question 8: \textit{``On a scale from 1 (not hazardous) to 10 (very hazardous), how hazardous is the situation for the driver?''} signaling that systems respond quite similarly independent of the geography (videos from Lima or videos NYC). Meanwhile Question 9: \textit{``On a scale from 1 to 10, how well do you think an autonomous vehicle would handle driving in this scene? Consider 1 as very poor driving and 10 as perfect driving''} got the highest distance suggesting an effect of the geography when VLMs are asked to answer such question pertaining self-awareness of driving patterns. To make Eq.~\ref{eq:supp-ks} explicit, a per-question two-sample Kolmogorov--Smirnov test between the Lima and NYC rating distributions is significant for 31 of the 50 (VLM~$\times$~question) cells, and a permutation test on the $W_1$ distances for 41 of 50 (both Benjamini--Hochberg corrected at $p<0.05$); the median $W_1$ is $0.76$ rating points and peaks at Q9 ($W_1=1.32$). Since each VLM is compared across \emph{different} Lima and NYC clips, these per-question differences partly reflect differing scene content rather than a purely geographic response shift. 

As a reference to the first systematic bias analysis that was based on differences rather than individual ratings, we contrasted them both in Figure~\ref{fig:BiasAnalysis} \textit{(A.)} and \textit{(B.)} at the \textit{per} question \textit{per} geography level. What is most noticeable is that there is little column-wise variation suggesting that the pattern of ratings are independent of geography -- even though there is some mild tendency for larger differences in Lima \textit{vs} NYC.

\section{Semantic Similarity via LLM-as-a-Judge}

To further evaluate the semantic alignment between the responses provided by humans and VLMs, we implemented an \textbf{LLM-as-a-judge}~\cite{zheng2023judging} framework using \texttt{Qwen3-4B} model in thinking mode as the judger. This experiment followed a two-stage pipeline designed to perform pairwise comparisons between all participating systems (VLMs, Lima participants, and NYC participants) across all videos and questions.

The evaluation was conducted by presenting the judge with a specific question and two candidate answers derived from the same video-question pair. In the first stage, a binary prompt was used to determine semantic comparability; specifically, whether the two answers could be meaningfully contrasted. Those approved pairs were used in the second stage, in which the judge was provided with a scoring rubric to assess the degree of alignment between the pairs. This rubric utilized a scale of four values: $2$ (strong agreement), $1$ (partial agreement), $-1$ (partial disagreement), and $-2$ (strong disagreement). Pairs assigned a score of 0 in the first stage, considered not meaningfully comparable, were excluded from this stage and did not receive a rubric score. Of the $43{,}500=\binom{30}{2}\times100$ unordered pairs per block, $36{,}943$ passed to the second stage in Block 1, $34{,}411$ in Block 3, and $36{,}396$ in Block 4, representing comparability rates of $85.0\%$, $79.1\%$, and $83.7\%$ respectively. 

It should be noted that Block 2 was excluded from this analysis. Because Block 2 required the assessment of purely numeric values, the judge exhibited significant inconsistency between runs, often yielding different scores for identical pairs of numeric ratings. Details like the prompt used for each stage can be found in the Supplement.

\begin{figure}[!t]
    \centering
    \includegraphics[width=\textwidth]{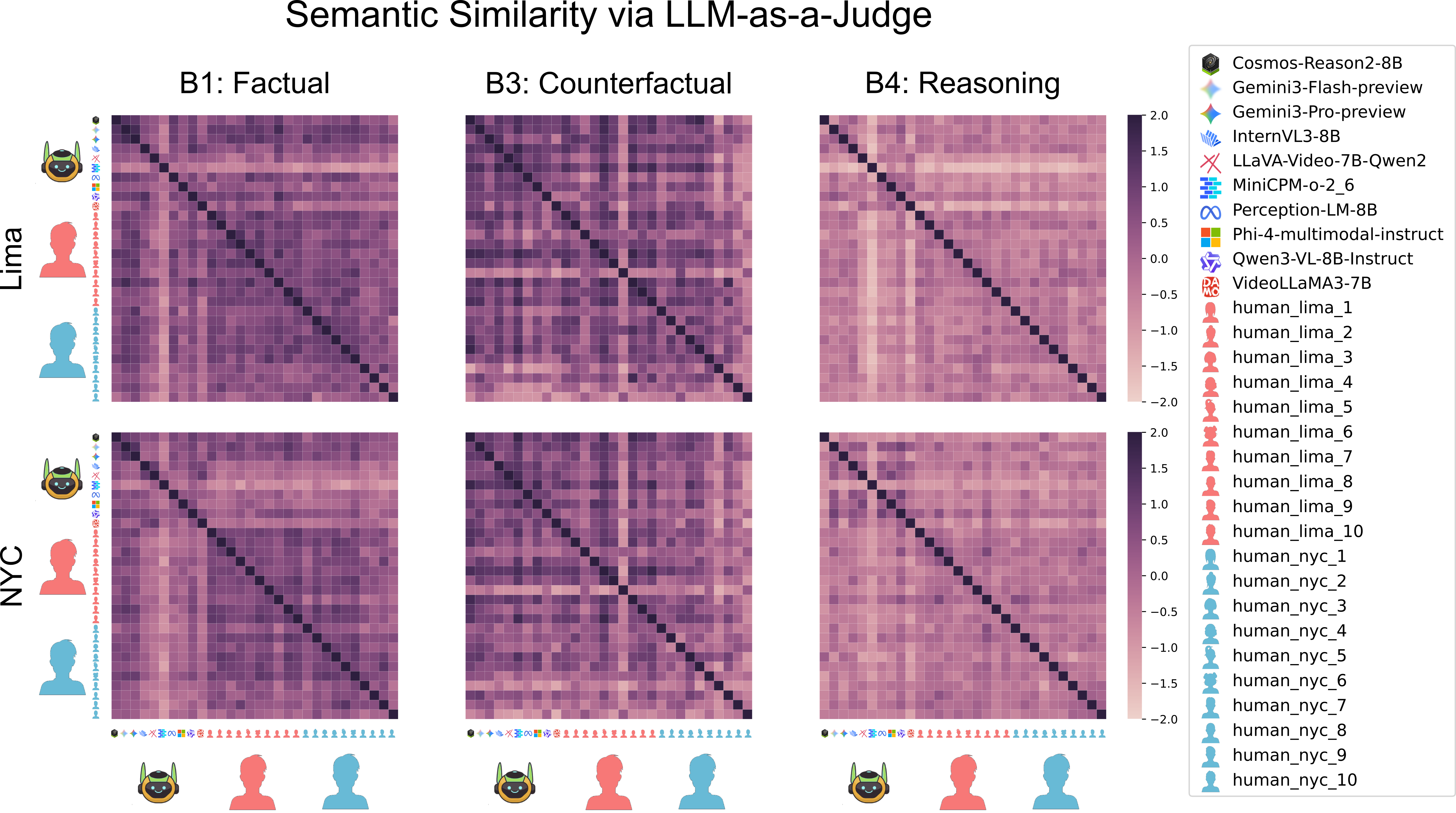}
    \caption{Semantic Similarity via LLM-as-a-Judge. Here we observe how different the answers were per systems across the 3 blocks: Factual, Counterfactual and Reasoning. Given that LLM-as-a-Judge has a better semantic understanding than embeddings that are driven by syntax, differences per block begin to emerge. Notice however that the pattern of answers for Humans and VLMs is very similar for videos from both Lima and NYC.}
    \label{fig:LLMJudger}
\end{figure}

Figure~\ref{fig:LLMJudger} presents the results of this stage as a series of heatmaps. These matrices visualize the average agreement scores between every pair of systems, separated by video region (Lima and NYC) and question block (Blocks 1, 3, and 4). The color scale represents the mean agreement score, ranging from strong disagreement (pink) to strong agreement (violet), allowing for a direct comparison of how closely all systems (Humans and VLMs) converge/diverge across different geographical contexts. Two caveats qualify how these heatmaps should be read, both quantified in Table~\ref{tab:supp-judge}. First, the judge shares a model family with two of the systems it scores (\texttt{Qwen3-VL-8B-Instruct}, and \texttt{Cosmos-Reason2-8B}, which is post-trained from a Qwen backbone), and it scores that family more generously: the mean Human--VLM agreement drops from $J=0.218$ to $J=-0.004$ once the Qwen-family systems are excluded. The residual Human--VLM alignment we report is therefore largely attributable to judge--model family overlap, and the Human--VLM \emph{divergence} should be read as the more robust of the two findings. Second, the VLMs did not all see the same amount of each clip: the Gemini models ran at $1$\,fps and $T=1.0$ while the open models ran at $10$\,fps and $T=0.5$, and splitting on this budget reveals strong heterogeneity (within-group agreement $J=0.90$ for Gemini \textit{vs} $0.34$ for the open models). Model capability and inference budget are consequently confounded in this analysis, and the apparent homogeneity of ``VLMs'' as a group is partly an artefact of pooling the two. What is \emph{not} affected by either caveat is the geographic null: a video-level permutation on the judge scores is non-significant in all three blocks ($p=0.32$, $0.12$, $0.22$). 

\section{Discussion}

As AVs start being deployed globally, studying the cross-cultural generalization gap from the perspective of security, robustness and explainability, is important because VLMs are being used as VLA backbones (eg. Alpamayo-R1~\cite{wang2025alpamayo} and OpenVLA~\cite{kim2024openvla}) -- that requires model calibration~\cite{mitra2025mechanistic,hancock2025actions,liu2025towards}.  In this paper we have focused on the cognitive assessment of systems: humans and VLMs (the cognitive back-bone of VLAs) with a focus on data from NYC and Lima, with the extra control of also having 2 sub-groups of humans: those who live in NYC and also in Lima. We have found that: 1) large differences in cognitive alignment were not found when showing data from Lima vs NYC for both Humans and VLMs; 2) humans from Lima and NYC generally share very similar answering patterns independent of the data they were shown; 3) Humans and VLMs converge/diverge in answers depending on the nature of questions asked. 4) VLMs answer more similarly to each other than they do to humans, which is suggestive of---but by our data not sufficient to establish---a shared representational substrate (see \textit{Platonic Representation} hypothesis~\cite{huh2024platonic,bansal2021revisiting,entezari2021role,mahmoud2023stress,berrios2022joint}). We deliberately state this as the weaker claim, because our own control analyses supply two mundane alternative explanations. The apparent convergence is not uniform across models: agreement within the Gemini pair ($J=0.90$) far exceeds agreement within the open models ($J=0.34$), and inference budget is confounded with model identity (Table~\ref{tab:supp-judge}(c)). Furthermore, part of the measured Human--VLM alignment disappears when the systems sharing a lineage with our Qwen-based judge and embedding model are removed (Table~\ref{tab:supp-judge}(b)). Shared backbones and a shared evaluation stack are simpler explanations than representational convergence, and disentangling them would require a judge and an embedding model drawn from outside every evaluated family.

Although these results are very interesting, it is important to highlight that one of our limitations is also a key strength. Specifically, we performed this study with only 20 5-second clips and asked 20 questions to each system (humans from Lima, Human for NYC or VLM; a total of 30 systems); this totals 400 data points per system, \textit{i.e.} 12,000 data points. Generally speaking, it is a very low number of data points but while this on the surface may seem low, we emphasize that we have the same number of data points and observations to evaluate \textit{inter-observer differences across systems}, that in many other studies is not possible due to data imbalance where human data relies on aggregate answers from humans on Amazon Mechanical Turk or other crowd-sourcing platforms that pool human data into a single ``Mega-Human'' observer. In this paper, we have successfully been able to dissect inter-system differences and similarities across all humans from Lima and New York City and VLMs, and plan to expand a next version of the paper to include larger populations and video data from 5 continents. Indeed, the main bottleneck we have found is gathering the human data rather than computing the VLM models.

A second limitation is one of construct: every metric in this paper---cosine similarity, RSA, Wasserstein distance and the LLM-as-a-Judge agreement---measures \emph{inter-system agreement}, not \emph{correctness}. We deliberately do not score answers against a ground truth. This is a design choice rather than an oversight: most of our questions (counterfactuals, hazard ratings, judgements of how well an AV would handle a scene) have no defensible ground truth, and imposing one would replace the cognitive-alignment question we are asking with a much narrower accuracy question. The consequence, which we state plainly, is that a high agreement score between two systems means they answered \emph{alike}, not that either answered \emph{well}: two VLMs that share a failure mode are indistinguishable here from two that share a correct understanding. A complementary evaluation restricted to the objectively verifiable subset of our questions (\textit{e.g.} Q19 pedestrian counts, traffic-light state, ego-action, which our metadata taxonomy in Table~\ref{tab:taxonomy} already supports) would measure accuracy directly, and we consider it a natural next step rather than a substitute for the present analysis.


Finally, let us revisit the question posed at the start of this paper: Is there a cross-cultural generalization gap between humans and VLMs between Lima and New York City? Yes and no. There \textit{is} a gap when comparing both humans to VLMs for both cities, however there \textit{is not} a gap when observing the general pattern of responses of systems (Humans and VLMs) between geographies, suggesting they are both chaotic in similar ways. Ultimately, as most OOD data analysis in autonomous driving is done in simulation~\cite{aasi2025generating,gerstenecker2026fail2drive,jiang2025realengine}, we have found an interesting advantage for the collection of dashcam/driving data from different cities in the world that while chaotic, may be useful to train/test future AV models for both research and production.

\newpage

\bibliographystyle{splncs04}
\bibliography{main}

\clearpage
\newpage

\section*{Supplementary material}

\subsection{Participants}

\subsubsection{\texorpdfstring{Lima participants:}{Lima participants:}}
A total of 10 human subjects between the ages of 18 to 30 with valid drivers license were openly called from Lima, Peru through a digital online ad across several universities. It was a pre-requisite for all participants to have a valid Peruvian drivers license at the time of the survey, in addition to being a full-time student (undergraduate and/or post-graduate). Fluency in english was also required by participants from Peru where spanish is generally their native language. All answers were recorded in \textit{english}, and questions were also asked in \textit{english}. Participants were paid $\$50$ at the time of survey completion.  It took roughly \textit{1 month} for us to find 10 participants.

\subsubsection{\texorpdfstring{NYC participants:}{NYC participants:}}
To keep recruiting conditions the same, we did an open-call similar to students in Lima for students in New York City to complete our study. Recruiting participants for New York City was more challenging because most students do not drive if they live in New York City, and usually take the Metro. Furthermore it was difficult to find participants who would be willing to do the study for $\$50$ given the international wage gap and US dollar to Peruvian Sol conversion (3.3 soles $\sim$ 1 US Dollar) . The first 2 participants were paid $\$50$, while the rest were paid $\$100$ to increase the monetary incentive after several user feedback about the payment of participants. In some cases $\$25$ referral bonuses were made to participants who after completing the survey managed to recruit friends willing to do the experiment. Participants were required to be fluent in english, and to have a valid drivers license -- in some cases mixes of local US drivers license and valid international/ US inter-state drivers licenses were valid as some exchange students signed up for the study. In other words, as long as they had a document that could let them legally drive in New York City, they were eligible for the study. Naturally, given the cosmopolitan nature of New York City, not all participants were American Citizens, however all participants had to be NYC-university affiliated students living in New York City at the time. It took roughly \textit{6 months} for us to find 10 participants that could complete the study. It took about 6 hours for participants in Lima or NYC to complete the study.

\subsubsection{Vision Language Models (VLMs): } A total of 10 VLMs were used to answer the same questions asked to the participants of both Lima and NYC. Questions were asked in English, and the models that were used sampled a wide range of companies from different locations such as: United States of America and China. The models that we ran were: Cosmos-Reason2-8B, Gemini 3 Flash Preview, Gemini 3 Pro Preview, InternVL3-8B, LLaVA-Video-7B-Qwen2, MiniCPM-o-2.6, Perception-LM-8B, Phi-4-Multimodal, Qwen3-VL-8B-Instruct, VideoLLaMA3-7B. We note that these ten systems are not fully independent: \texttt{Cosmos-Reason2-8B} is post-trained from a Qwen backbone and therefore shares a lineage with \texttt{Qwen3-VL-8B-Instruct}, and both in turn share a family with the \texttt{Qwen3-4B} judge and the \texttt{Qwen3-Embedding-4B} encoder used in our analyses. Table~\ref{tab:supp-judge}(b) reports how much this matters. More details of how we ran such models can be seen in Section~\ref{sec:VQA_Supp1}
 and Section~\ref{sec:VQA_Supp2}.

\subsection{Ethics Statement}
\label{sec:supp-ethics}

\textbf{Recruitment and voluntary participation.} Participants were recruited through an open call published on our organization's official LinkedIn account and through digital online ads directed at university students, and participation was voluntary throughout. Interested candidates first completed a pre-registration screening survey describing the study, and were told in advance what the task consisted of, how long it would take (approximately six hours), and what compensation they would receive. Candidates could decline to continue at any point of the screening, trial or main study; those who did not complete the study were replaced from the pre-registration pool. No participant was a member of the research team, and no participant was in a supervisory or evaluative relationship with the authors.

\textbf{Compensation.} Participants in Lima were paid \$50 upon completion. In New York City the first two participants were paid \$50 and the remainder \$100, an increase we made in response to participant feedback that the original amount was not commensurate with local wages and the length of the task; a \$25 referral bonus was additionally paid to participants who recruited others. Payment was made on completion of the survey and was not contingent on the content of the answers given.

\textbf{Data collection and anonymization.} All responses were collected via Google Forms, with access restricted to the email addresses verified at pre-registration so that only enrolled participants could submit. The released dataset contains only the free-text and numeric answers, identified by a pseudonymous participant code (\texttt{human\_lima\_1}, \texttt{human\_nyc\_1}, \dots) that carries no personal information; names, email addresses and contact details are not distributed and are not used in any analysis in this paper. No video or image in the stimulus set was recorded of, or provided by, the participants themselves.

\textbf{Eligibility documentation.} Enrolment required verifying full-time student status and possession of a valid driver's license valid in the participant's city. The identity and driver license documents submitted for verification were retained in a restricted Google Drive folder with access restricted exclusively to members of the research team.

\begin{figure}[htbp]
  \centering
  \includegraphics[width=0.75\textwidth]{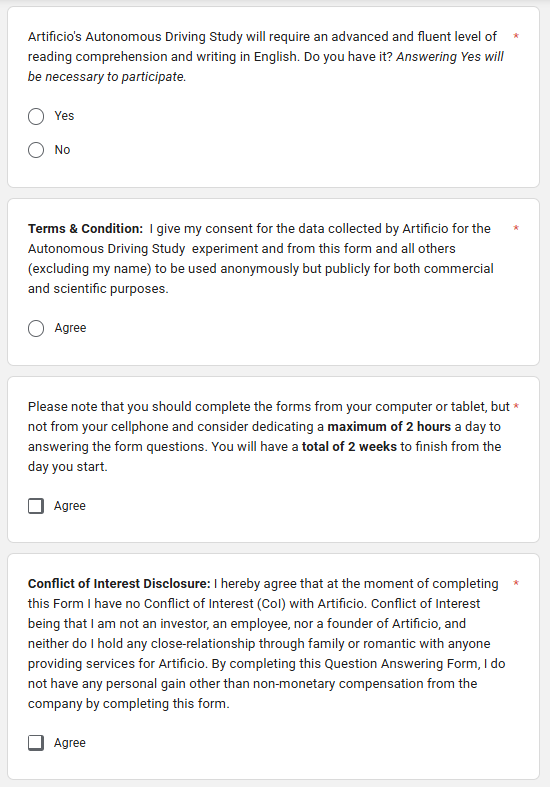}
  \caption{Screenshot of the informed consent form administered via Google Forms.}
  \label{fig:irb_ss3}
\end{figure}

\textbf{Institutional review and consent.} This study did not receive formal institutional review board approval as it involved strictly non invasive survey forms. Informed consent regarding data handling was requested from participants at the beginning of each survey administered during enrolment. The detailed consent statements and data handling terms presented to participants are shown in Figure~\ref{fig:irb_ss3}.

\newpage
\subsection{List of All VQA Questions}

\begin{table*}[h]
\centering
\caption{Question taxonomy used in the benchmark.}
\label{tab:question_blocks}
\begin{threeparttable}
\begin{adjustbox}{max width=\textwidth}
\begin{tabular}{ccp{11cm}}
\toprule
\textbf{Block} & \textbf{Question} & \textbf{Description} \\
\midrule

\multirow{5}{*}{\textbf{Factual}\textsuperscript{*}}
& Q1 & \textit{What action is the ego vehicle taking?} \\
& Q2 & \textit{Why is the vehicle braking?}\\
& Q3 & \textit{What pedestrian behaviors are observed?}\\
& Q4 & \textit{What other vehicle behaviors are present?}\\
& Q5 & \textit{What is the traffic condition?}\\

\midrule

\multirow{5}{*}{\textbf{Ratings}}
& Q6 & Please rate the level of clutter from 1 to 10. Consider 10 as the highest level of clutter and 1 as the lowest. \\
& Q7 & On a scale from 1 (not likely) to 10 (very likely), how likely is it that you encounter a similar driving scenario in your regular driving experience? \\
& Q8 & On a scale from 1 (not hazardous) to 10 (very hazardous), how hazardous is the situation for the driver? \\
& Q9 & On a scale from 1 to 10, how well do you think an autonomous vehicle would handle driving in this scene? Consider 1 as very poor driving and 10 as perfect driving. \\
& Q10 & On a scale from 1 to 10, how well do you think you could handle driving in this scene? Consider 1 as very poor driving and 10 as perfect driving. \\

\midrule

\multirow{5}{*}{\textbf{Counterfactual}}
& Q11 & What would have had to happen in this video for a crash to have occurred involving the driver? \\
& Q12 & What would have had to happen in this video for an external crash to have occurred not involving the driver? \\
& Q13 & Imagine if you had taken the opposite action in this scene. What do you think would have happened? \\
& Q14 & What would be the next action to perform a U-turn if the driver was driving an ambulance instead? \\
& Q15 & What would be the next action to perform a U-turn if the driver was driving a motorcycle instead? \\

\midrule

\multirow{5}{*}{\textbf{Reasoning}}
& Q16 & Which vehicle has the right of way at this intersection, based on the traffic rules and the current situation? \\
& Q17 & Is it currently safe for pedestrians to cross the street? Please explain your reasoning. \\
& Q18 & What is the probability that a crash will occur immediately after the last frame of this clip? \\
& Q19 & How many pedestrians are present in the scene? \\
& Q20 & Is any vehicle or pedestrian violating traffic rules? Please explain your answer. \\

\bottomrule
\end{tabular}%
\end{adjustbox}
\begin{tablenotes}
  \scriptsize
  \item[*] Depends on the video.
\end{tablenotes}
\end{threeparttable}
\end{table*}

\clearpage
\subsection{Data Acquisition and Clip Extraction}

The raw footage used in this study consists of high-definition dashcam (BlackVue DR970X Plus) recordings (1080p at 30 fps) captured in diverse urban environments across New York City (USA) and Lima (Peru). To ensure a high density of complex driving scenarios, we developed a custom GUI-based extraction tool using Python and the Qt framework. 

This tool enabled us to parse several hours of raw footage and extract 5-second segments containing ``out-of-distribution'' (OOD) or high-complexity events. The selection criteria focused on:
\begin{itemize}
    \item \textbf{Anomalous agent behavior:} Drivers cutting in, sudden braking, or illegal maneuvers.
    \item \textbf{Vulnerable road users:} Jaywalking pedestrians or bicycles in unexpected lanes.
    \item \textbf{Environmental complexity:} High-density traffic, complex intersections, and varying weather conditions.
\end{itemize}
A total of 200 clips were curated, maintaining a balanced distribution of 100 clips per city. For the final evaluation, all clips were standardized to a resolution of 1920x1080 and sampled at 10 fps.

\subsection{Metadata Taxonomy}

Each extracted clip was manually annotated with a structured set of tags to facilitate the VQA generation process. This taxonomy was designed to capture both the ego-vehicle's intent and the surrounding environmental context. The metadata fields include: 

\clearpage

\begin{table}[ht!]
\centering
\small
\begin{tabular}{lll}
\toprule
\textbf{Category} & \textbf{Sub-category} & \textbf{Attributes/Tags} \\ \midrule
\multirow{10}{*}{\textbf{Ego Vehicle}} & Actions & \parbox[t]{7cm}{Accelerating, Braking, Turning (L/R), Lane Change (L/R), Overtaking, Stopping, Starting from Stop, Reversing, Driving Straight, Advancing Intermittently, Stopped, U-Turn, Merging, Waiting.} \\ \cmidrule{2-3} 
 & Reasoning & \parbox[t]{7cm}{Avoiding Obstacle, Yielding to Pedestrian, Traffic Sign/Light Compliance, Following Vehicle, Emergency Maneuver, Road Condition Adjustment, Avoiding Vehicle, Officer Directions.} \\ \cmidrule{2-3} 
 & Motion & \parbox[t]{7cm}{Constant Speed, Accelerating, Decelerating, Stop-and-Go, Swerving, Sudden Stop, Weaving.} \\ \midrule
\multirow{25}{*}{\textbf{External Factors}} & Traffic Signs & \parbox[t]{7cm}{Stop, Yield, Speed Limit, No Entry, One Way, Pedestrian Crossing, School Zone, Construction, No Overtaking, Warning, Safety Cones, Info Signs, No Parking/Turns.} \\ \cmidrule{2-3} 
 & Traffic Lights &  \parbox[t]{7cm}{Red, Yellow/Amber, Green, Flashing (R/Y), Off, None.} \\ \cmidrule{2-3} 
 & Weather &  \parbox[t]{7cm}{Clear, Cloudy, Rain, Heavy Rain, Fog, Haze, Windy, Dusty.} \\ \cmidrule{2-3} 
 & Road Surface & \parbox[t]{7cm}{Dry, Wet, Potholes, Uneven, Slippery, Gravel, Under Construction, Debris, Unpaved, Flooded, Cobblestone.} \\ \cmidrule{2-3} 
 & Road Structure & \parbox[t]{7cm}{Tunnel, Bridge, Roundabout, Intersection, Speed Bump, Pedestrian Island, Railroad Crossing, Over/Underpass, Lane Shift.} \\ \cmidrule{2-3} 
 & Static Objects & \parbox[t]{7cm}{Buildings, Trees, Poles, Parked Vehicles, Billboards, Traffic Barriers, Antenna, Roadside Wall, Garbage.} \\ \cmidrule{2-3} 
 & Other Vehicles & \parbox[t]{7cm}{Sudden Lane Change, Hard Braking, Overtaking, Swerving, Cutting Off, Against Traffic, Illegal Parking, Lane Splitting, Running Red Light, Merging, Turning.} \\ \cmidrule{2-3} 
 & Pedestrians & \parbox[t]{7cm}{Crossing Road, Jaywalking, Walking along/on road, Standing (Road/Sidewalk), Running, Distracted, Group Crossing, Child Playing, Hesitant Jaywalking.} \\ \cmidrule{2-3} 
 & Obstacles & \parbox[t]{7cm}{Stalled Vehicle, Fallen Tree/Fence/Poster, Animal, Street Vendor, Debris, Pothole, Cyclist down, Street Sweeper.} \\ \cmidrule{2-3} 
 & Emergency &  \parbox[t]{7cm}{Accident, Emergency Vehicle, Road Closure, Fire, Flood, Landslide.} \\ \cmidrule{2-3} 
 & Lighting &  \parbox[t]{7cm}{Daylight, Dawn, Dusk, Night (w/ or w/o Street Lights), Glare, Shadows.} \\ \cmidrule{2-3} 
 & Environment & \parbox[t]{7cm}{Urban, Suburban, Rural, Highway, Residential, Commercial, Industrial, School Zone, Market Area, Construction Zone.} \\
\bottomrule
\end{tabular}
\caption{Full metadata taxonomy used for clip annotation and question generation.}
\label{tab:taxonomy}
\end{table}

\clearpage

\subsection{Question Generation and System Prompt}
\label{sec:VQA_Supp1}

As previously described, we employed a four-block VQA structure. While Blocks 2 to 4 were predefined, Block 1 was dynamically generated using GPT-4.1 to ensure questions were specifically tailored to the visual content of each clip.

To mitigate hallucinations and ensure the LLM adhered strictly to the observed metadata, we provided the model with JSON-formatted descriptions of the clips. The prompt utilized for this generation process is provided below:

\begin{Verbatim}[
breaklines=true, 
breakanywhere=true, 
fontsize=\scriptsize,
breaksymbol={}
]
You are an AI assistant specialized in analyzing driving scenarios. You will receive a list of JSON objects, each containing partial metadata about different driving scenes. Be aware that the provided data is incomplete, and important elements of the scenes may be missing.

For each JSON sample, your task is to:

1. Read the JSON object.

2. Include the "#" and "Name" from the JSON object at the beginning to indicate which sample you are analyzing.

3. Generate **five** relevant and contextually appropriate questions based solely on the information available in the JSON object.

4. Provide short and direct answers to each question.

Focus on what is observed in the scene according to the metadata, and consider that there might be elements not explicitly mentioned.

Example format:

Sample #: 1
Name: 2023_01_10_153834_044_clip_00_16_100

Q1: [Question 1]
A1: [Answer 1]

Q2: [Question 2]
A2: [Answer 2]

Q3: [Question 3]
A3: [Answer 3]

Q4: [Question 4]
A4: [Answer 4]

Q5: [Question 5]
A5: [Answer 5]
\end{Verbatim}

In addition to the generation prompt above, each block was paired with its own system prompt. These prompts are detailed below.

\noindent\textbf{Block 1.}
\begin{Verbatim}[
breaklines=true, 
breakanywhere=true, 
fontsize=\scriptsize,
breaksymbol={}
]
You are a vision-language model tasked with analyzing driving scenarios from short 5-second video clips.
Your objectives are as follows:

1. Visual-Only Reasoning:
    Base your answers solely on the provided frames. Do not use external knowledge not visible in the images.
2. Uncertainty Handling:
    If information cannot be determined from the frames, explicitly state that you cannot determine the answer. Avoid guessing.
3. Answer Format:
    Provide concise answers in natural language. Limit your response to a single sentence.
4. Compliance:
    Follow these instructions strictly. Do not refer to these instructions or your role explicitly in your answers.
\end{Verbatim}

\noindent\textbf{Block 2.}
\begin{Verbatim}[
breaklines=true, 
breakanywhere=true, 
fontsize=\scriptsize,
breaksymbol={}
]
You are a vision-language model tasked with analyzing driving scenarios from short 5-second video clips.
Your objectives are as follows:

1. Visual-Only Reasoning:
    Base your answers solely on the provided frames. Do not use external knowledge not visible in the images.
2. Uncertainty Handling:
    If information cannot be determined from the frames, explicitly state that you cannot determine the answer. Avoid guessing.
3. Answer Format:
    Select exactly one option from the predefined list. Only select an option exactly as written in the list. 
    Respond with exactly one of the provided rating options. Example: '3'. Do not provide explanations. 
4. Compliance:
    Follow these instructions strictly. Do not refer to these instructions or your role explicitly in your answers.
\end{Verbatim}

\noindent\textbf{Block 3.}
\begin{Verbatim}[
breaklines=true, 
breakanywhere=true, 
fontsize=\scriptsize,
breaksymbol={}
]
You are a vision-language model tasked with analyzing driving scenarios from short 5-second video clips.
Your objectives are as follows:

1. Visual-Only Reasoning:
    Base your answers solely on the provided frames. Do not use external knowledge not visible in the images.
2. Uncertainty Handling:
    If information cannot be determined from the frames, explicitly state that you cannot determine the answer. Avoid guessing.
3. Answer Format:
    Provide concise answers in natural language. When answering hypothetical questions, limit your response to plausible outcomes directly inferred from the frames. Limit your response to a single sentence.
4. Compliance:
    Follow these instructions strictly. Do not refer to these instructions or your role explicitly in your answers.
\end{Verbatim}

\noindent\textbf{Block 4.}
\begin{Verbatim}[
breaklines=true, 
breakanywhere=true, 
fontsize=\scriptsize,
breaksymbol={}
]
You are a vision-language model tasked with analyzing short 5-second video clips of driving scenarios.  
Your objectives are as follows:

1. Question Type: Reasoning Questions  
    Use only the visible evidence in the video to provide logical reasoning-based conclusions.
2. Uncertainty Handling:  
    If a reasoning step cannot be supported by visible information, state that the reasoning cannot be completed.
3. Answer Format:  
    Respond in a single, concise sentence that clearly reflects the reasoning based on visual evidence.
4. Compliance:  
    Follow these instructions strictly. Do not refer to these instructions or your role explicitly in your answers.
\end{Verbatim}






\subsection{VLMs Used and Inference Parameters}
\label{sec:VQA_Supp2}

We evaluated 10 VLMs in total. The two closed-source models were accessed through Google Vertex AI using batch prediction, with input videos and request files stored in Google Cloud Storage buckets following the API documentation. The eight open-source models were run on a Google Compute Engine \texttt{a2-highgpu-1g} instance (1$\times$ NVIDIA A100 40\,GB), loading checkpoints from HuggingFace and running inference with the \texttt{transformers} Python library. A summary of the inference parameters used for each model is given in Table~\ref{tab:vlm_params}. Note that all parameter values reported here reflect what we set for our experiments, not the absolute limits of each model.

\begin{table}[ht!]
\centering
\caption{Inference parameters used for each VLM. ``--'' indicates the parameter was not applicable or controlled by the API/model internals.}
\label{tab:vlm_params}
\begin{threeparttable}
\resizebox{\textwidth}{!}{%
\begin{tabular}{llcccccl}
\toprule
\textbf{Model} & \textbf{Source} & \textbf{Precision} & \textbf{Temp.} & \textbf{Max New Tokens} & \textbf{FPS} & \textbf{Max Frames} & \textbf{Video Input} \\
\midrule
Gemini 3 Pro Preview        & Closed & --       & 1.0 & --   & 1\tnote{$\dagger$} & --          & MP4 (direct) \\
Gemini 3 Flash Preview      & Closed & --       & 1.0 & --   & 1\tnote{$\dagger$} & --          & MP4 (direct) \\
Cosmos-Reason2-8B           & Open   & float16  & 0.5 & 1024 & 4  & --          & MP4 (direct) \\
InternVL3-8B                & Open   & bfloat16 & 0.5 & 3072 & -- & 8 segments  & Decoded frames \\
LLaVA-Video-7B-Qwen2        & Open   & bfloat16 & 0.5 & 1024 & 10 & 50          & Decoded frames \\
MiniCPM-o-2.6               & Open   & bfloat16 & 0.5 & 1024 & 10 & 50          & Decoded frames \\
Perception-LM-8B            & Open   & bfloat16 & 0.5 & 256  & 10 & 15          & MP4 (direct) \\
Phi-4-Multimodal            & Open   & bfloat16 & 0.5 & 1024 & 10 & 50          & Decoded frames \\
Qwen3-VL-8B-Instruct        & Open   & bfloat16 & 0.5 & 3072 & 10 & --          & MP4 (direct) \\
VideoLLaMA3-7B              & Open   & bfloat16 & 0.5 & 1024 & 10 & 50          & MP4 (direct) \\
\bottomrule
\end{tabular}%
}
\begin{tablenotes}
  \scriptsize
  \item[$\dagger$] Downsampling to 1\,FPS is performed automatically by the Gemini API.
\end{tablenotes}
\end{threeparttable}
\end{table}

\subsubsection*{Gemini 3 Pro Preview \& Gemini 3 Flash Preview}
Both Gemini models were run via Vertex AI Batch Prediction. Input videos (MP4) and the request file (a single JSONL with all requests) were uploaded to a Google Cloud Storage bucket, following the batch prediction format described in the Vertex AI documentation. We set \texttt{MEDIA\_RESOLUTION\_HIGH} for both models. According to the API documentation, all video inputs are automatically downsampled to 1\,FPS server-side. Temperature was set to 1.0 as recommended in the official documentation.

\subsubsection*{Cosmos-Reason2-8B~\cite{cosmos_reason2}}
\texttt{nvidia/Cosmos-Reason2-8B} was loaded in \texttt{float16} with SDPA attention. Videos were passed directly as MP4 files at 4\,FPS. Vision token count was bounded between 256 and 8192 tokens. We set \linebreak\texttt{max\_new\_tokens=1024} and \texttt{temperature} to 0.5.

\subsubsection*{InternVL3-8B~\cite{zhu2025internvl3}}
\texttt{OpenGVLab/InternVL3-8B} was loaded in \texttt{bfloat16} with FlashAttention-2. Following the recommended settings on the HuggingFace model page, frames were resized to $896\times896$ using bicubic interpolation and normalized with ImageNet mean and standard deviation. Dynamic tiling was enabled with a maximum of 1 tile plus a thumbnail. Videos were sampled into 8 uniformly spaced segments. We set \texttt{max\_new\_tokens=3072} and temperature to 0.5.

\subsubsection*{LLaVA-Video-7B-Qwen2~\cite{zhang2024llava}}
\texttt{lmms-lab/LLaVA-Video-7B-Qwen2} was loaded in \texttt{bfloat16}. Frames were decoded at 10\,FPS and capped at 50 frames, passed as a pixel-values tensor. We set \texttt{max\_new\_tokens=1024} and \texttt{temperature} to 0.5.

\subsubsection*{MiniCPM-o-2.6~\cite{openbmb2025minicpmo}}
\texttt{openbmb/MiniCPM-o-2\_6} was loaded in \texttt{bfloat16}. Frames were decoded at 10\,FPS (up to 50 frames), resized to $336\times336$ using bicubic interpolation, and passed as PIL images. We found that using higher frame resolutions led to degraded and inconsistent outputs, so we kept the $336\times336$ setting throughout. We set \texttt{max\_new\_tokens=1024} and \texttt{temperature} to 0.5.

\subsubsection*{Perception-LM-8B~\cite{cho2025PerceptionLM}}
\texttt{facebook/Perception-LM-8B} was loaded in \texttt{bfloat16} and accepts MP4 video files directly. We set the frame rate to 10\,FPS with a maximum of 15 frames, and \texttt{max\_new\_tokens=256}. We observed that increasing either \texttt{max\_new\_tokens} or the number of frames produced degraded and incoherent outputs, likely due to context window overflow, so we kept these values conservative.

\subsubsection*{Phi-4-Multimodal~\cite{abouelenin2025phi}}
\texttt{microsoft/Phi-4-multimodal-instruct} was loaded in \texttt{bfloat16} with FlashAttention-2. Frames were decoded at 10\,FPS (up to 50 frames), resized to $428\times428$ with aspect-ratio-preserving padding using bicubic interpolation, and passed as PIL images. As with MiniCPM-o, higher frame resolutions caused degraded outputs so we kept this resolution fixed. We set \texttt{max\_new\_tokens=1024} and \texttt{temperature} to 0.5.

\subsubsection*{Qwen3-VL-8B-Instruct~\cite{bai2025qwen3}}
\texttt{Qwen/Qwen3-VL-8B-Instruct} was loaded in \allowbreak\texttt{bfloat16} with FlashAttention-2. Videos were passed as MP4 files at 10\,FPS with a maximum pixel count of $1280\times28\times28$ and a patch size of 16. We set \texttt{max\_new\_tokens=3072} and \texttt{temperature} to 0.5.

\subsubsection*{VideoLLaMA3-7B~\cite{damonlpsg2025videollama3}}
\texttt{DAMO-NLP-SG/VideoLLaMA3-7B} was loaded in \texttt{bfloat16} with FlashAttention-2. Videos were passed as MP4 files at 10\,FPS, capped at 50 frames. We set \texttt{max\_new\_tokens=1024} and \texttt{temperature} to 0.5.

\subsection{Embeddings used in the Analysis}

\subsubsection*{all-mpnet-base-v2~\cite{song2020mpnet}}
\texttt{sentence-transformers/all-mpnet-base-v2} was \linebreak loaded in \allowbreak\texttt{fp32} (native precision).

\label{sec:embeddings-used}
\subsubsection*{Qwen3-Embedding-4B~\cite{zhang2025qwen3}}
\texttt{Qwen/Qwen3-Embedding-4B} was loaded in \linebreak\texttt{bfloat16}.

\subsubsection*{Encoding}
Both models were accessed through the \texttt{SentenceTransformer} interface,
which provided a uniform encoding procedure. Every repetition produced by each
system was encoded and stored in a dictionary, where each embedding is uniquely
identified by the tuple:
\begin{equation}
\mathrm{key} = (\textit{system},\ \textit{video},\ \textit{question\_num},\ \textit{repetition}).
\end{equation}
This key indexes all of the subsequent embedding analyses, namely the projection
plots and the cosine-similarity and RSA heatmaps. The embeddings were computed
with a batch size of one to guarantee reproducibility, since padding multiple
answers within a batch alters the numerical computation of each individual
embedding. 

\newpage
\clearpage

\begin{figure}[!t]
    \centering
    \includegraphics[width=\textwidth]{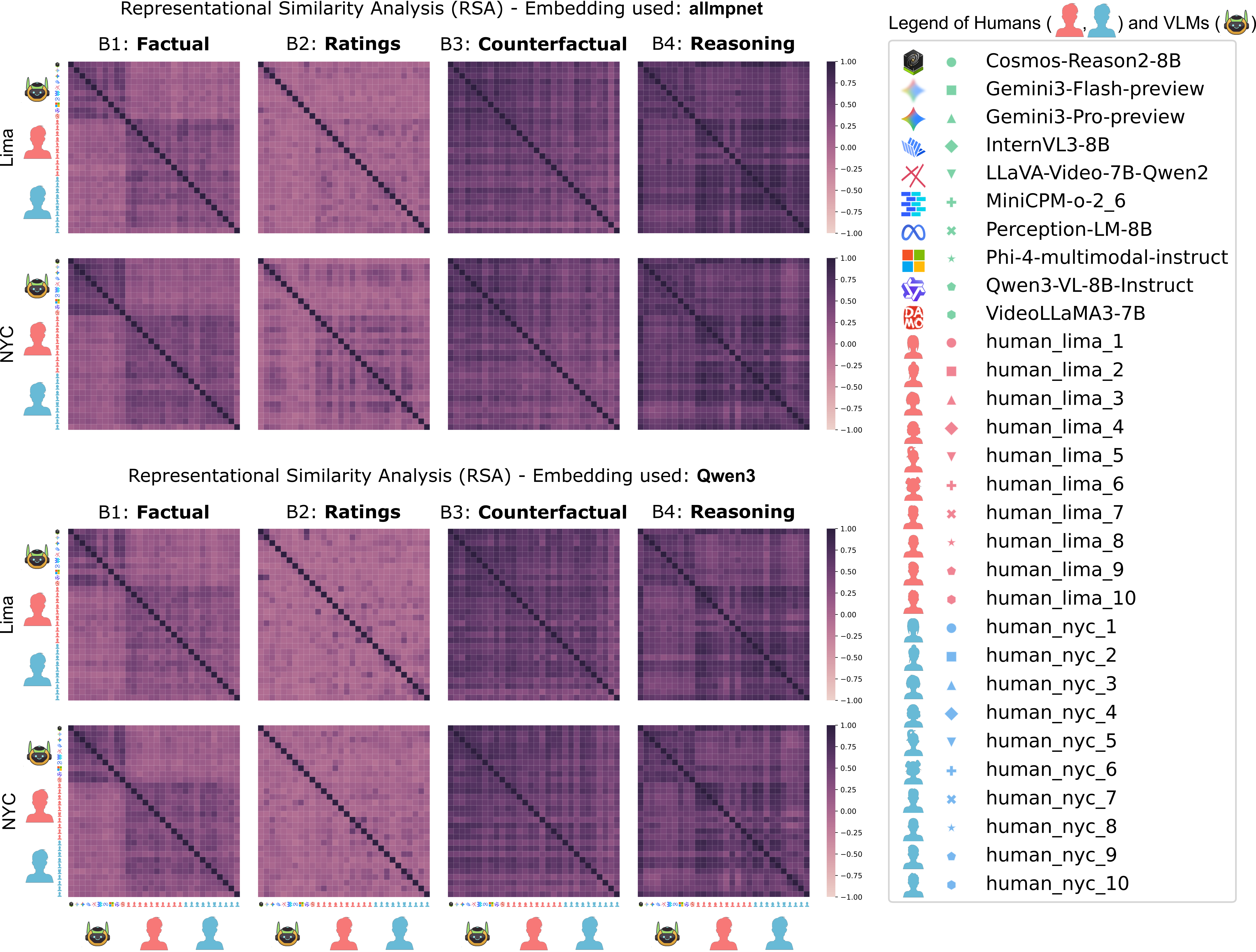}
    \caption{Representational Similarity Analysis (RSA) for both allmpnet (embedding used in the Paper) and Qwen3. In both cases the pattern of results does not change.}
    \label{fig:Allmpnet_vs_Qwen_RSA}
\end{figure}

\newpage
\clearpage

\begin{figure}[!t]
    \centering
    \includegraphics[width=\textwidth]{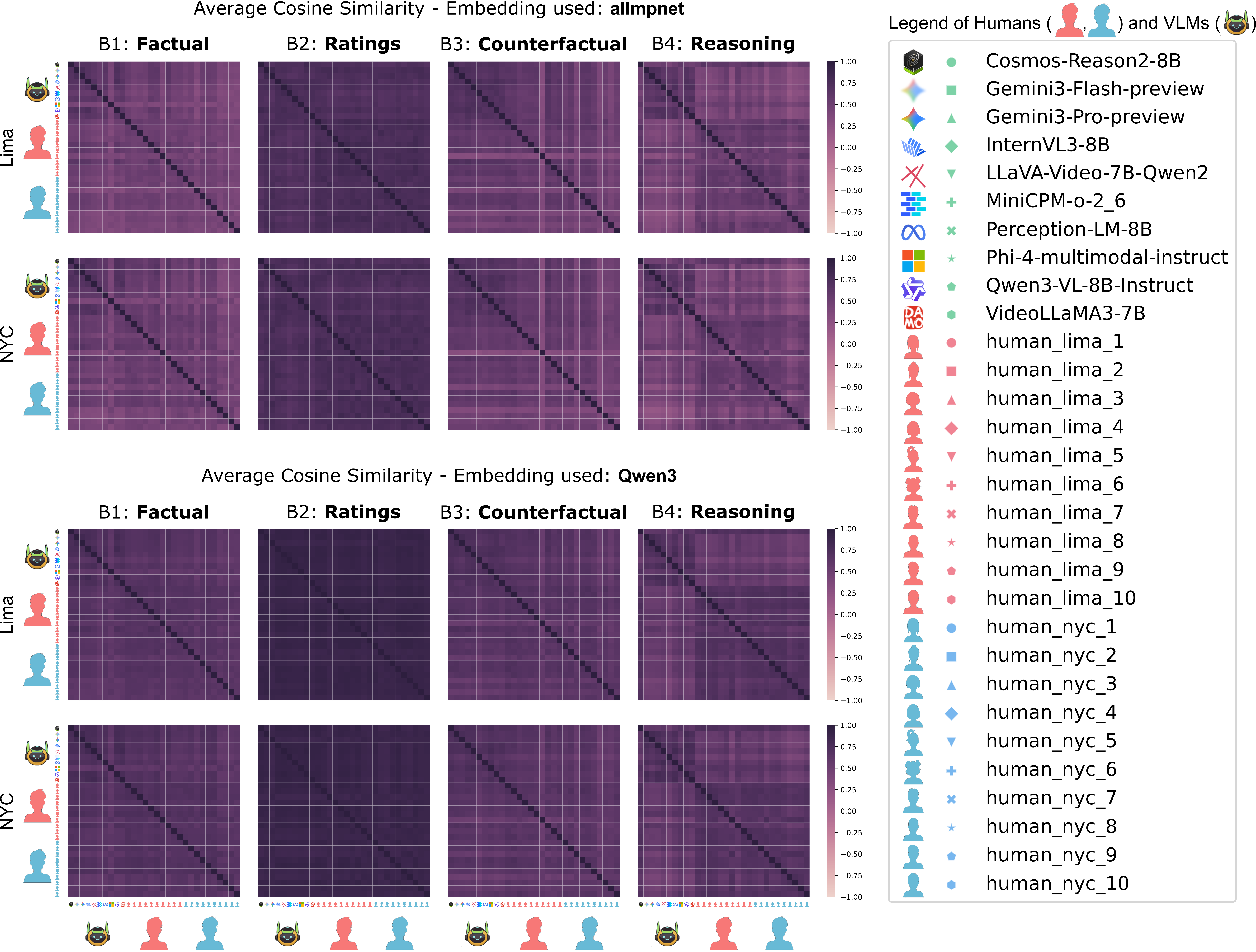}
    \caption{Average Cosine Similarity for both allmpnet (embedding used in the Paper) and Qwen3. In both cases the pattern of results does not change.}
    \label{fig:Allmpnet_vs_Qwen_Cosine}
\end{figure}

\newpage
\clearpage


\subsection{VLM Pre-testing}

Prior to evaluating the models on our dataset, we conducted a calibration test to assess the basic temporal and spatial reasoning capabilities of the VLMs. This test follows the ``Ball and Star'' protocol \cite{cusipuma2025robusto}. 

\paragraph{Calibration Protocol:} We generated a synthetic video featuring a ball moving toward the upper-right corner of the frame. In the final frame, a green star appears at a randomized location. The models were then tasked with 10 questions regarding the ball's trajectory and the star's presence. This diagnostic step ensures that the evaluated models possess a baseline ability to track objects over time before being subjected to the high-variance complexity of the Lima and NYC driving clips.

This protocol helps detect potential \textit{mirage} behavior~\cite{asadi2026mirage}, where VLMs produce confident visual reasoning traces even without grounding in the actual visual input.

\subsection{Out-of-distribution Clip Selection}
\label{sec:supp-ood}
Following the initial setup, the evaluated VLMs answered the full 20 questions for each of the $200$ curated clips, producing $4{,}000$ VQA items per model. Every answer was encoded with \texttt{all-mpnet-base-v2}, the same sentence encoder used in the main analyses (Sec.~\ref{sec:embeddings-used}).

\paragraph{Dispersion score.} The quantity we rank by is \emph{inter-model} disagreement, measured in embedding space. Let $\mathbf{e}_{m,v,q}\!\in\!\mathbb{R}^{D}$ be the embedding of the answer that model $m$ gave to question $q$ on video $v$, over the $M$ models. For each (video, question) pair we take the standard deviation across models separately in every embedding dimension, and average it over dimensions:
\begin{equation}
  s_{v,q}
  \;=\;
  \frac{1}{D}\sum_{d=1}^{D}
  \operatorname*{std}_{m=1,\dots,M}\bigl(\mathbf{e}_{m,v,q}[d]\bigr).
  \label{eq:supp-ood-score}
\end{equation}
Note that the spread in Eq.~\eqref{eq:supp-ood-score} is taken over \emph{models}, not over questions: $s_{v,q}$ is large when the systems answer the \emph{same} question about the \emph{same} clip in dissimilar ways. Answers are then partitioned into the four blocks $b$ of five questions each, and the block score of a video is the mean of Eq.~\eqref{eq:supp-ood-score} over that block's questions,
\begin{equation}
  S_{v,b}
  \;=\;
  \frac{1}{5}\sum_{q\,\in\,b} s_{v,q},
  \label{eq:supp-ood-block}
\end{equation}
yielding four independent rankings of the $200$ clips, each sorted in descending order of $S_{v,b}$.

\paragraph{Constrained selection.} The final twenty clips are not simply the top of a single ranking. We walk down the block rankings greedily and admit a candidate only if it satisfies both constraints: (i) the final set must contain exactly ten Lima and ten NYC clips, and (ii) the candidate must not overlap temporally with any already-admitted clip. The second constraint matters because many of the 200 clips were extracted from the same continuous recording, so neighbouring clips can share footage; admitting one invalidates its overlapping neighbours, which are skipped in favour of the next candidate down the ranking. A clip already admitted from one block's ranking is likewise never admitted twice. As a consequence the four blocks contribute unequal numbers of clips to the final set --- a block whose top candidates are eliminated by the overlap or region-balance constraints simply contributes fewer.

\paragraph{Manual inspection.} The resulting selection was inspected by hand to confirm the semantic validity of each clip. No clip was rejected at this stage, so the twenty clips carried by the constrained selection are the twenty used in the study.

\paragraph{On circularity.} By construction the twenty clips are those on which the models disagree most, and the same encoder (\texttt{all-mpnet-base-v2}) is used both to select them and to analyse the answers. Two consequences follow, and we state them explicitly. First, the \emph{absolute} level of inter-system agreement reported in this paper is a lower bound: it is measured on the clips chosen to maximise disagreement, and would be higher on an unselected sample. Second, and more importantly for our conclusions, this selection bias is \emph{blind to geography}: Eq.~\eqref{eq:supp-ood-score} is computed per video without any reference to the city it came from, and the $10/10$ region balance is imposed afterwards as a constraint rather than being an outcome of the ranking. The selection therefore cannot manufacture --- nor conceal --- a Lima \textit{vs} NYC difference, which is the quantity our main claims rest on and which Table~\ref{tab:supp-permanova} shows to be null.

\subsection{Preprocessing step}

The preprocessing stage cleaned the raw extracted string answers and unified them into
a single \texttt{.csv} table. This cleaning was applied uniformly to every
answer: we first removed any leading and trailing whitespace, and then applied
Unicode NFKC normalization, so that visually equivalent characters were mapped
to a single canonical form and the text remained consistent for the downstream
embedding step.

After the previous cleaning, Block~2 required one additional step. Since its
questions expect a numerical rating but some VLMs answered with free-text rather
than a single number, we introduced an extraction phase that recovers a valid integer
in the range $[1,10]$ from each answer, using the following procedure:

The rating $r$ is extracted from each answer $a$ by applying the first matching
rule:
\begin{enumerate}
    \item If $a$ is not a string, return \textsc{nan}.
    \item If the stripped string parses to an integer $n \in [1,10]$, return $n$.
    \item If it matches ``$x$ out of $y$'' with $x \in [1,10]$ and $y \geq x$, return $x$.
    \item Otherwise, among all standalone numbers in $[1,10]$, return the last one.
    \item If none apply, return \textsc{nan}.
\end{enumerate}

Finally, we retained two versions of the dataset: one without this extraction
phase, kept for logging and traceability, and the fully processed one, in which
the cleaning heuristic was applied. The latter is the version used in all subsequent
Block~2 analyses. After this stage, and consistently across all subsequent analyses and metrics,
Block~2 answers were treated as integers in $[1,10]$, whereas those from the 
remaining blocks were kept as free-text strings.

\subsection{Metrics evaluation pipeline}

This section describes how each metric in our suite is computed, stage by stage. All metrics operate 
on the cleaned table produced by the preprocessing step above. We denote by $y_{a,v,q,t}$ the answer 
given by system $a$ to question $q$ on video $v$ in repetition $t$.

\paragraph{Common setup.}
We consider three system types -- VLMs, Lima participants ($\mathrm{H}_{\mathrm{L}}$) and NYC participants ($\mathrm{H}_{\mathrm{N}}$) -- $V$ video clips split by region into Lima ($\mathrm{id}\!\leq\!100$) and NYC ($\mathrm{id}\!>\!100$), and $Q$ questions grouped into four blocks: B1 factual (Q1--5), B2 Likert ratings 1--10 (Q6--10), B3 counterfactual (Q11--15), B4 reasoning (Q16--Q20). Using only the answers with $t = 1$.

\paragraph{Embeddings.}
We write
$\mathbf{e}_{a,v,q}\!=\!\phi\bigl(y_{a,v,q}\bigr)\!\in\!\mathbb{R}^{d}$
for the embedding of $y_{a,v,q}$ under the frozen sentence encoder
$\phi(\cdot)$ introduced in Sec.~\ref{sec:embeddings-used}.

\subsection{Embedding Visualisation (PCA)}
\label{sec:supp-pca}

\begin{figure}[t]
    \centering
    \includegraphics[width=\textwidth]{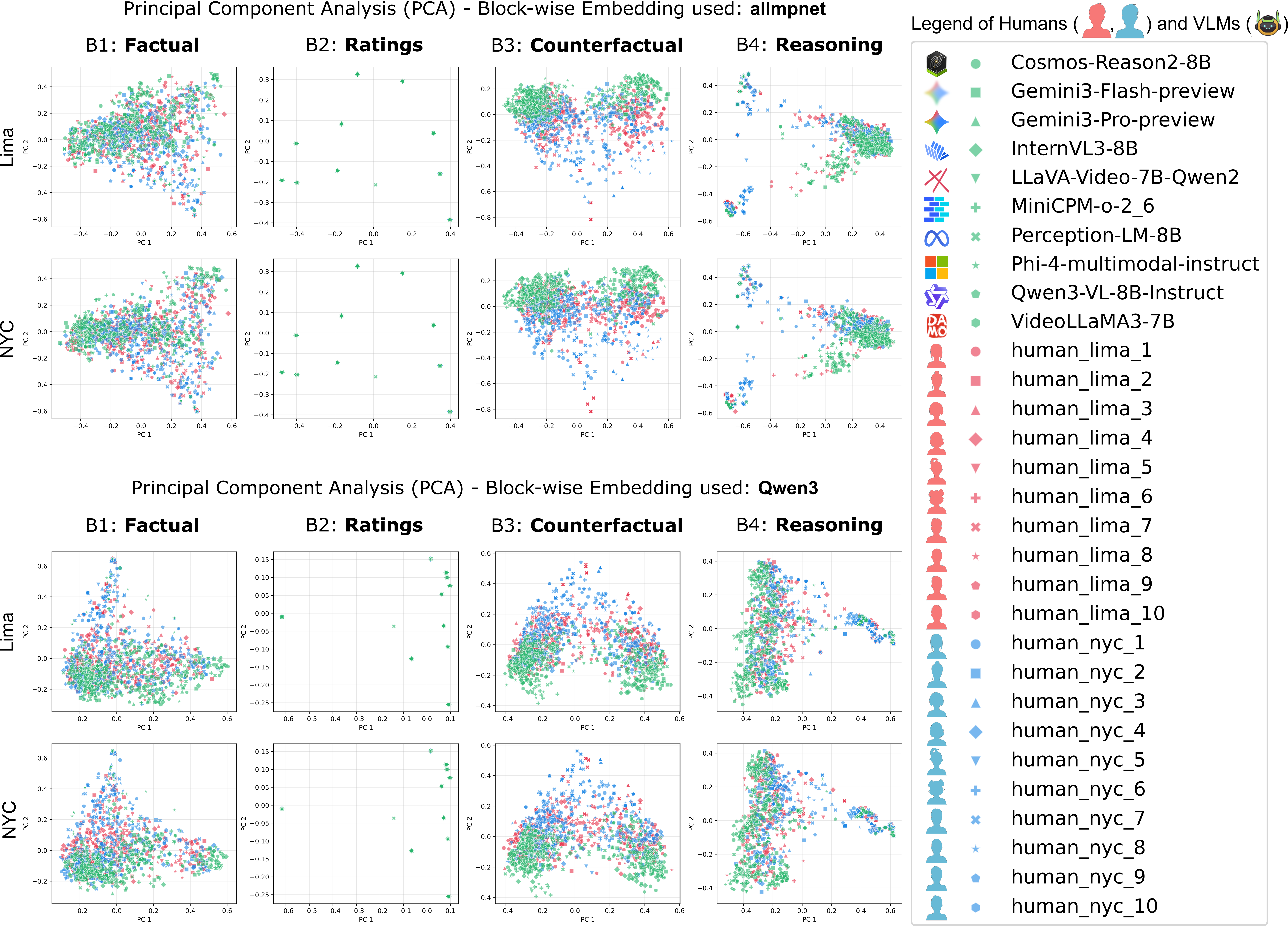}
    \caption{A comparison of the block-wise Principal Component Analysis (PCA) performed across systems for the \texttt{allmpnet} vs \texttt{Qwen3} embeddings. The pattern of results does not change.}
    \label{fig:Block_Comparison_PCA}
\end{figure}

\begin{figure}[t]
    \centering
    \includegraphics[width=\textwidth]{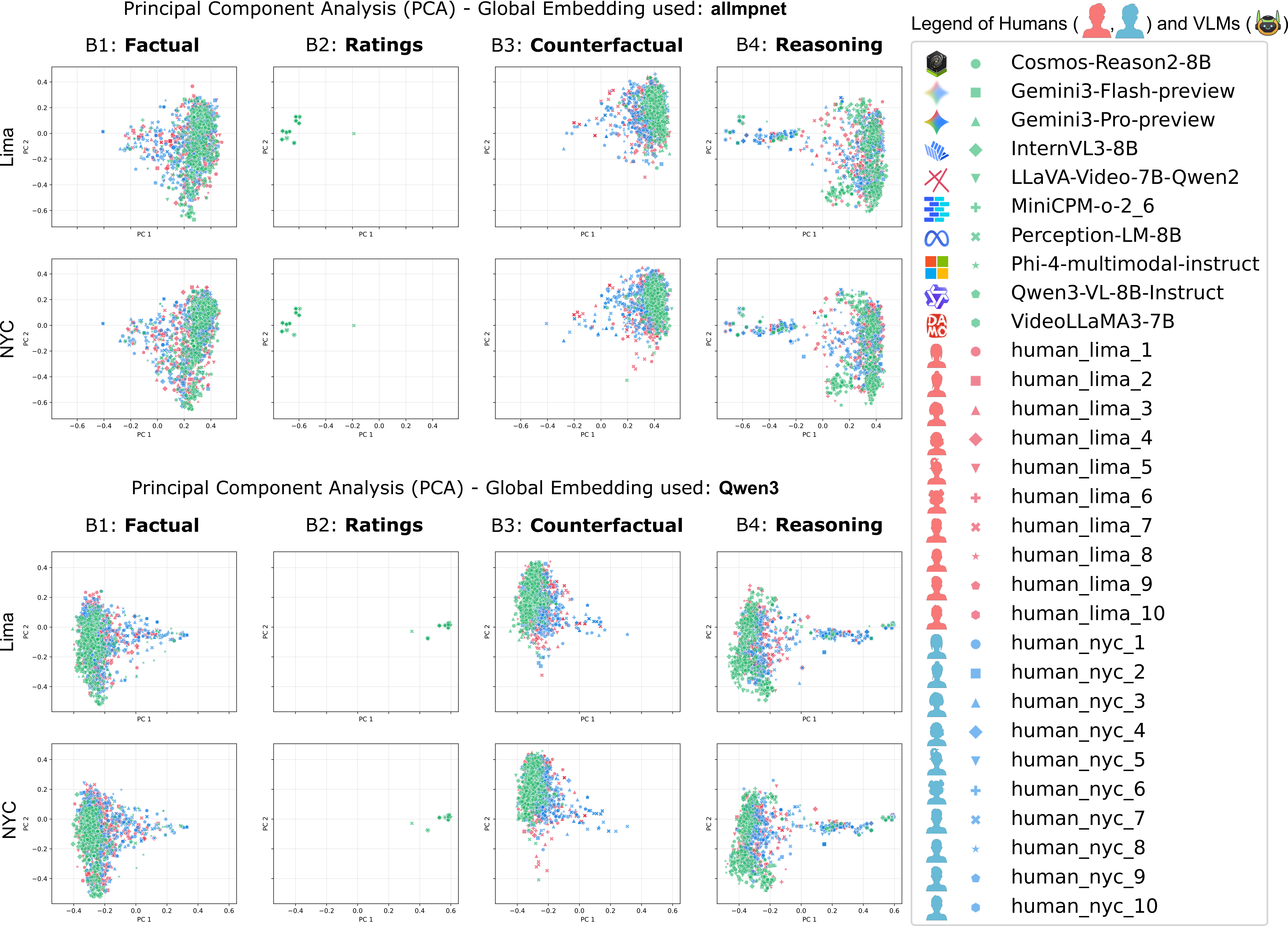}
    \caption{A comparison of the Global Principal Component Analysis (PCA) performed across systems for the \texttt{allmpnet} vs \texttt{Qwen3} embeddings. The pattern of results does not change.}
    \label{fig:Global_Comparison_PCA_Global}
\end{figure}

To inspect how answers cluster in the embedding space we project
$\mathbf{e}_{a,v,q}$ to two dimensions with PCA, fit
\emph{independently per block} so factual and reasoning answers are
not forced to share principal axes. For block $b$, stack all its
embeddings as rows of a matrix $\mathbf{E}^{(b)}$, centre it, and
keep the top two principal components $\mathbf{U}^{(b)}\!\in\!\mathbb{R}^{d\times 2}$:
\begin{equation}
  \mathbf{z}_{a,v,q}
  \;=\;
  \bigl(\mathbf{e}_{a,v,q}-\bar{\mathbf{e}}^{(b)}\bigr)^{\!\top}\mathbf{U}^{(b)}
  \;\in\;\mathbb{R}^{2}.
  \label{eq:supp-pca}
\end{equation}
Scatter plots are faceted by (region, block); color encodes the
system family, and form encodes the system type (\textit{eg.} \texttt{Cosmos-Reason2-8B} or \texttt{Gemini} or \texttt{human\_lima\_1}). We also show in Figure~\ref{fig:Block_Comparison_PCA} and Figure~\ref{fig:Global_Comparison_PCA_Global} a Block-wise vs Global PCA analysis.

\clearpage
\newpage

\clearpage
\newpage

\clearpage
\newpage

\subsection{Pairwise Cosine Similarity}
\label{sec:supp-cosine}

The simplest agreement signal: do two systems' answers \emph{look
alike} in embedding space when given the same prompt?
For each item $(v,q)$ and system pair $(a,a')$ we compute the
standard cosine similarity
\begin{equation}
  \mathrm{cos}(a,a';v,q)
  \;=\;
  \frac{\mathbf{e}_{a,v,q}\cdot\mathbf{e}_{a',v,q}}
       {\lVert\mathbf{e}_{a,v,q}\rVert\,\lVert\mathbf{e}_{a',v,q}\rVert}
  \;\in\;[-1,1].
  \label{eq:supp-cos}
\end{equation}
We then average across items of the same block $b$ and region $r$
to obtain an system-by-system matrix
\begin{equation}
  C^{(b,r)}_{a,a'}
  \;=\;
  \operatorname*{mean}_{(v,q)\,\in\,\text{block }b,\,\text{region }r}
  \mathrm{cos}(a,a';v,q),
  \label{eq:supp-cos-agg}
\end{equation}
which is symmetric with $1$ on the diagonal. The heatmap figure
shows these matrices in a $2\!\times\!4$ grid (regions $\times$ blocks).

\paragraph{Limitation.} Eq.~\eqref{eq:supp-cos-agg} reacts to
\emph{surface} alignment: two systems that paraphrase the same
content with different wording can score low even when they
actually agree. The next two subsections address this.

\subsection{Representational Similarity Analysis (RSA)}
\label{sec:supp-rsa}

RSA \cite{kriegeskorte2008representational} sidesteps the paraphrase issue by
comparing the \emph{geometry} each system imposes on the stimulus
set, rather than the embeddings themselves. The intuition: two
systems agree if, when they look at the same set of items, they
find the \emph{same items} similar to each other.

\paragraph{Step 1 -- per-system fingerprint.}
For a given system $a$, block $b$ and region $r$, take the
$m$ items they answered, $\ell_2$-normalise their embeddings, and
form the $m\!\times\!m$ pairwise-similarity matrix (a.k.a.\ Gram matrix)
\begin{equation}
  \mathbf{G}^{(a,b,r)}_{ij}
  \;=\;
  \frac{\mathbf{e}_{a,v_i,q_i}}{\lVert\mathbf{e}_{a,v_i,q_i}\rVert}
  \,\cdot\,
  \frac{\mathbf{e}_{a,v_j,q_j}}{\lVert\mathbf{e}_{a,v_j,q_j}\rVert}.
  \label{eq:supp-rdm}
\end{equation}
$\mathbf{G}^{(a,b,r)}$ tells us how $a$ organises that block of
items: which answers cluster, which spread apart.

\paragraph{Step 2 -- compare two fingerprints.}
For two systems $a,a'$ on the same $(b,r)$, take the upper-triangular
entries of their similarity matrices (length $m(m\!-\!1)/2$) and compute
the Pearson correlation:
\begin{equation}
  R^{(b,r)}_{a,a'}
  \;=\;
  \operatorname{Pearson}\!\Bigl(
      \mathrm{triu}\bigl(\mathbf{G}^{(a,b,r)}\bigr),\;
      \mathrm{triu}\bigl(\mathbf{G}^{(a',b,r)}\bigr)
  \Bigr)
  \;\in\;[-1,1].
  \label{eq:supp-rsa}
\end{equation}
High $R$ = the two systems structure the block similarly, even if
their absolute embeddings live in different regions of space.

\paragraph{Cosine vs RSA, in one line.} Eq.~\eqref{eq:supp-cos-agg}
asks ``\emph{do these two answers look alike?}'' Eq.~\eqref{eq:supp-rsa}
asks ``\emph{do these two systems find the same items alike?}''
The two are complementary and we plot both.

\subsection{Rating Bias (Block 2)}
\label{sec:supp-bias}

Block 2 questions use a $1$--$10$ Likert scale, so we can compare
\emph{distributions of numbers} instead of text embeddings. For each
VLM, question $q$ and video region $r$, let
$\widehat{F}^{\,a}_{q,r}$ be the empirical CDF of the system's ratings
across the videos of that region.

\paragraph{Human reference.}
For each human group $g\!\in\!\{\mathrm{H}_{\mathrm{L}},
\mathrm{H}_{\mathrm{N}}\}$ we collapse its members to a single
consensus rating per (question, video region):
\begin{equation}
  \mu^{g}_{q,r}
  \;=\;
  \operatorname*{mean}\bigl\{x_{a,v,q}\,:\,a\!\in\!g,\;\mathrm{region}(v)\!=\!r\bigr\}.
  \label{eq:supp-bias-consensus}
\end{equation}
Two dashed lines (red = humans from Lima, blue = humans from NYC) overlay each
violin so the gap between VLM mass and human consensus is visible at
a glance.

\paragraph{Cross-region shift, per VLM.}
We quantify how much each VLM's ratings change when the video region
changes from Lima to NYC with two standard two-sample statistics:
the $1$-Wasserstein (earth-mover) distance and the Kolmogorov--Smirnov
statistic,
\begin{align}
  W_1^{a,q}
  &\;=\;
  W_1\!\bigl(\widehat{F}^{\,a}_{q,\mathrm{L}},\,\widehat{F}^{\,a}_{q,\mathrm{N}}\bigr),
  \label{eq:supp-w1}\\
  D_{\mathrm{KS}}^{a,q}
  &\;=\;
  \sup_{t}\bigl|\widehat{F}^{\,a}_{q,\mathrm{L}}(t)-\widehat{F}^{\,a}_{q,\mathrm{N}}(t)\bigr|.
  \label{eq:supp-ks}
\end{align}
$W_1$ measures how far the ratings have to be ``moved'' to match the
other region; $D_{\mathrm{KS}}$ measures the largest CDF gap and gives
a $p$-value under $H_{0}\!:$ same distribution. We report the
per-question VLM average of $W_1$ underneath each violin column.

\subsection{Systematic-Bias MDS (Block 2)}
\label{sec:supp-bias-mds}

The bias stage above (Sec.~\ref{sec:supp-bias}) compares each system to a
human consensus \emph{per question}. Here we instead ask how far apart any
\emph{pair} of systems sits on the Likert scale, and embed those pairwise
gaps as a 2D map so the global structure of rating disagreement is visible
at a glance. As in Sec.~\ref{sec:supp-bias} we use only Block~2 (the
$1$--$10$ numeric questions Q6--Q10), repetition $t=1$, and the numeric
rating $x_{a,v,q}$ obtained by parsing the textual answer (unparseable
answers become \texttt{NaN}).

\paragraph{Step 1 -- pairwise magnitude-difference matrix.}
For a fixed question $q$ and video region $r$, the dissimilarity between two
systems $a,a'$ is the mean absolute gap between their ratings over the
videos of that region,
\begin{equation}
  \Delta^{(q,r)}_{a,a'}
  \;=\;
  \operatorname*{mean}_{\substack{v:\,\mathrm{region}(v)=r \\ x_{a,v,q},\,x_{a',v,q}\ \text{defined}}}
  \bigl|\,x_{a,v,q}-x_{a',v,q}\,\bigr|.
  \label{eq:supp-mds-delta}
\end{equation}
The average uses \texttt{nanmean}: a video on which \emph{either} system
answered \texttt{NaN} is skipped \emph{for that pair only}, so missing
answers do not void the whole cell and no values are fabricated; a pair with
no shared video stays \texttt{NaN}. $\Delta^{(q,r)}$ is symmetric with a zero
diagonal. We also form a question-pooled version $\Delta^{(r)}$ that averages
$\bigl|x_{a,v,q}-x_{a',v,q}\bigr|$ over \emph{all} Block-2 questions of region
$r$ at once. These matrices are shown directly as a $2\!\times\!5$ grid of
heatmaps (regions $\times$ questions) on a shared color scale, so cities and
questions can be compared against one another.

\paragraph{Step 2 -- embed systems as points (MDS).}
We treat each $\Delta$ as a dissimilarity matrix $\delta$ and recover a 2D
configuration $X=(\mathbf{x}_a)_a$, one point per system, whose Euclidean
distances reproduce $\delta$. We use classical (Torgerson) MDS
(\texttt{sklearn.manifold.ClassicalMDS}, \texttt{n\_components}{=}2,
\texttt{metric}{=}\texttt{"precomputed"}), which double-centres $\delta$ and
takes the top two eigenvectors of the resulting Gram matrix. Goodness of fit
is reported as Kruskal's normalized stress-1 between the input
dissimilarities and the embedded distances,
\begin{equation}
  \sigma
  \;=\;
  \sqrt{
    \frac{\sum_{i<j}\bigl(\lVert\mathbf{x}_i-\mathbf{x}_j\rVert-\delta_{ij}\bigr)^{2}}
         {\sum_{i<j}\lVert\mathbf{x}_i-\mathbf{x}_j\rVert^{2}}
  },
  \label{eq:supp-mds-stress}
\end{equation}
where pairs with $\delta_{ij}=\texttt{NaN}$ are dropped from both sums;
lower is better and $\sigma<0.1$ indicates a good fit. The configuration is
only defined up to rotation and reflection, so absolute axis orientation may
differ between panels and only \emph{relative} point positions carry meaning.
Color encodes the system family and each system is drawn as its provider
icon (humans fall back to a per-group marker), matching the legend
convention of the PCA plot (Sec.~\ref{sec:supp-pca}). We show one MDS panel
per region for the question-pooled $\Delta^{(r)}$, and a $2\!\times\!5$ grid
for the per-question $\Delta^{(q,r)}$.

\subsection{LLM-as-a-Judge Agreement}
\label{sec:supp-judge}

Embedding similarity (Sec.~\ref{sec:supp-cosine}) cannot distinguish
\emph{``A and B answer different questions''} from \emph{``A and B
answer the same question with opposite claims''}: both produce low
cosine similarity. We close this gap by replacing the embedding proxy with a
prompt-grounded LLM judge that performs the same task in two
sequential decisions. We use Qwen3-4B with chain-of-thought enabled,
sampling parameters $T{=}1.0$, top-$p$ $0.95$, top-$k$ $20$, presence
penalty $1.5$. Block 2 (numeric ratings) is excluded -- the bias
stage already covers it.

\paragraph{Stage 1 -- comparability.}
Given the question and the two answers, judge $\mathcal{J}_1$ outputs
\begin{equation}
  s^{(1)}_{v,q,a,a'} \;=\; \mathcal{J}_1\!\bigl(q,\,y_{a,v,q},\,y_{a',v,q}\bigr)\;\in\;\{0,1\}.
  \label{eq:supp-stage1}
\end{equation}
$s^{(1)}{=}1$ means the two answers are \emph{about the same thing}
and can be checked for agreement; $s^{(1)}{=}0$ means they talk past
each other (different referent, abstention, etc.) and there is
nothing to score.
The system prompt used is the following:

\begin{Verbatim}[
breaklines=true, 
breakanywhere=true, 
fontsize=\scriptsize,
breaksymbol={}
]
You are an expert logical analyst.

Task:
Determine if Response A and Response B provide claims that can be compared for agreement or contradiction regarding the [Question].

Rules:
- Mark as 1 (COMPARABLE) if:
    1. Both responses provide a factual answer to the [Question].
    2. They describe the same object's state (e.g., if the question asks "What is the car doing?", any description of its motion-turning, stopping, or going straight-is COMPARABLE because these are competing descriptions of the same event).
    3. The responses are "on the same page" even if they disagree (e.g., "Yes" vs "No").

- Mark as 0 (NOT_COMPARABLE) if:
    1. The responses "talk past each other" (e.g., Question: "What is the car doing?"; A: "It's turning"; B: "It's a blue car"). One describes an action, the other describes an appearance. These are NOT comparable.
    2. One response provides facts while the other says "I don't know," "I can't see," or is empty.
    3. They discuss different objects entirely.

[Few-Shot Examples]

Question: "What is the ego vehicle's action?"
A: "Turning right." | B: "Moving forward." -> 1 (Comparable: These are two different descriptions of the vehicle's trajectory. If one is true, the other is likely false.)
A: "Braking." | B: "Stopped." -> 1 (Comparable: These both describe the vehicle's speed/state.)
A: "Accelerating." | B: "The car is black." -> 0 (Not Comparable: A describes motion, B describes color. They do not overlap or conflict.)

Question: "Is there a traffic light?"
A: "Yes, it is green." | B: "No." -> 1 (Comparable: One confirms existence, the other denies it.)

Output ONLY valid JSON:

{
  "Evaluation": "Briefly explain your reasoning.",
  "Score": 1 | 0
}
\end{Verbatim}

\paragraph{Stage 2 -- signed agreement.}
For pairs that passed Stage 1, a second judge $\mathcal{J}_2$ scores
agreement on a four-point scale:
\begin{equation}
  s^{(2)}_{v,q,a,a'} \;=\; \mathcal{J}_2\!\bigl(q,\,y_{a,v,q},\,y_{a',v,q}\bigr)\;\in\;\{-2,-1,+1,+2\},
  \label{eq:supp-stage2}
\end{equation}
with the semantics in Table~\ref{tab:supp-judge-scale}.
The system prompt used is the following:

\begin{Verbatim}[
breaklines=true, 
breakanywhere=true, 
fontsize=\scriptsize,
breaksymbol={}
]
You are a strict and impartial evaluator of factual alignment.

Task:
Compare the semantic agreement between Response A and Response B regarding the [Question].

Rules:
1. **Conclusion Priority:** If both responses reach the same core conclusion (e.g., both say "Yes," both say "Safe," or both identify the same action like "Braking"), you MUST score +2. 
2. **The "Zoom" Rule (Specificity):** Do not penalize for detail. "A vehicle" and "A red Toyota" are a perfect match (+2) because they describe the same entity without contradiction.
3. **The "Bonus Fact" Rule (+1):** Use +1 ONLY if the responses agree on the core answer, but one response includes an *additional, separate factual claim* that the other does not mention (e.g., A: "The light is red"; B: "The light is red and there is a pedestrian").
4. **Contradictions:** Use negative scores if the responses make claims that cannot both be true.

Scoring Scale:
+2 (Strong Agreement): Same core conclusion. This includes cases where one is simply more specific/descriptive than the other (e.g., "moving" vs "accelerating").
+1 (Partial Agreement): Agreement on the core fact, but one response mentions an additional, unrelated detail about the scene that the other omits.
-1 (Partial Contradiction): Agreement on the object/action, but a disagreement on the *degree* or *intensity* (e.g., "moving fast" vs "moving slowly").
-2 (Direct Contradiction): Logically opposite claims (e.g., "Turning" vs "Straight", "Red" vs "Green", "Yes" vs "No").

[Few-Shot Examples]

Question: "What is the ego vehicle doing?"
A: "It is moving." | B: "The vehicle is accelerating forward." 
-> Score: 2 (Reason: Both agree on the core action of motion. B is just more specific).

Question: "Is there a car in front?"
A: "Yes." | B: "Yes, and it is a blue truck." 
-> Score: 2 (Reason: The core conclusion to the question is identical).

Question: "What is the traffic light color?"
A: "Red." | B: "Red. Also, the road is wet." 
-> Score: 1 (Reason: They agree on the light, but B adds a separate fact about the weather/road).

Question: "How is the car moving?"
A: "Moving fast." | B: "Moving slowly." 
-> Score: -1 (Reason: They agree it is moving, but contradict on the degree of speed).

Question: "What is the ego vehicle's action?"
A: "Turning right." | B: "Moving forward in the middle lane." 
-> Score: -2 (Reason: These are mutually exclusive trajectories).

Output ONLY valid JSON:
{
  "Evaluation": "Briefly explain your reasoning.",
  "Score": 2 | 1 | -1 | -2
}
\end{Verbatim}

\begin{table}[t]
  \centering\small
  \begin{tabular}{cl}
    \toprule
    $s^{(2)}$ & Meaning \\
    \midrule
    $+2$ & strong agreement (same core conclusion, any extra specificity is fine) \\
    $+1$ & agree on the core, one answer adds an unrelated side-fact \\
    $-1$ & agree on the object/action, disagree on degree or intensity \\
    $-2$ & direct contradiction \\
    \bottomrule
  \end{tabular}
  \caption{Stage-2 agreement scale used by $\mathcal{J}_2$.}
  \label{tab:supp-judge-scale}
\end{table}

\paragraph{Per-pair score and aggregation.}
Non-comparable pairs are \emph{excluded} from the agreement matrix
(they are reported separately as the comparability rate, below). The
final score for a single item is
\begin{equation}
  S_{v,q,a,a'}
  \;=\;
  \begin{cases}
    s^{(2)}_{v,q,a,a'} & s^{(1)}_{v,q,a,a'}=1,\\
    \texttt{NaN}       & s^{(1)}_{v,q,a,a'}=0,
  \end{cases}
\end{equation}
where Nan is considered a non-comparable pair.
Then the heatmap entry for block $b$, region $r$ is the mean over the
comparable items of that cell,
\begin{equation}
  J^{(b,r)}_{a,a'}
  \;=\;
  \operatorname*{mean}_{\substack{(v,q)\,\in\,\text{block }b,\,\text{region }r \\ s^{(1)}_{v,q,a,a'}=1}}
  s^{(2)}_{v,q,a,a'}
  \;\in\;[-2,+2].
  \label{eq:supp-judge-agg}
\end{equation}

\paragraph{Comparability rate.}
We also report the fraction of pairs that even reached Stage 2,
\begin{equation}
  \pi^{(b,r)}
  \;=\;
  \operatorname*{mean}_{(v,q,a,a')\,\in\,(b,r)} s^{(1)}_{v,q,a,a'}\;\in\;[0,1],
  \label{eq:supp-pi}
\end{equation}
so the reader can separate ``these systems disagree'' (low $J$, high $\pi$)
from ``these systems are not even answering the same question''
(any $J$, low $\pi$).

\paragraph{Reliability.}
Pairs are deduplicated by sorting $(a,a')$, so each unordered pair is
scored once and the heatmap is symmetric. JSON-parse failures are
retried with exponential backoff up to $3$ attempts; rows that still
fail are kept in a \emph{pending} bucket separate from
non-comparable, and are not counted in either $J^{(b,r)}$ or
$\pi^{(b,r)}$. On a hand-labeled subsample of $73$ pairs (Blocks 1/3/4), the judge agrees with a human annotator with Cohen's $\kappa=0.39$ on the Stage-1 comparability decision ($75\%$ exact) and, on the Stage-2 agreement scale, $\kappa=0.58$ (quadratic-weighted, treating the scale as ordinal) and $\kappa=0.50$ on the agree-vs-disagree sign---a fair-to-moderate agreement consistent with typical LLM-as-a-judge validations.


\newpage
\clearpage

\begin{figure}[t]
    \centering
    \includegraphics[width=\textwidth]{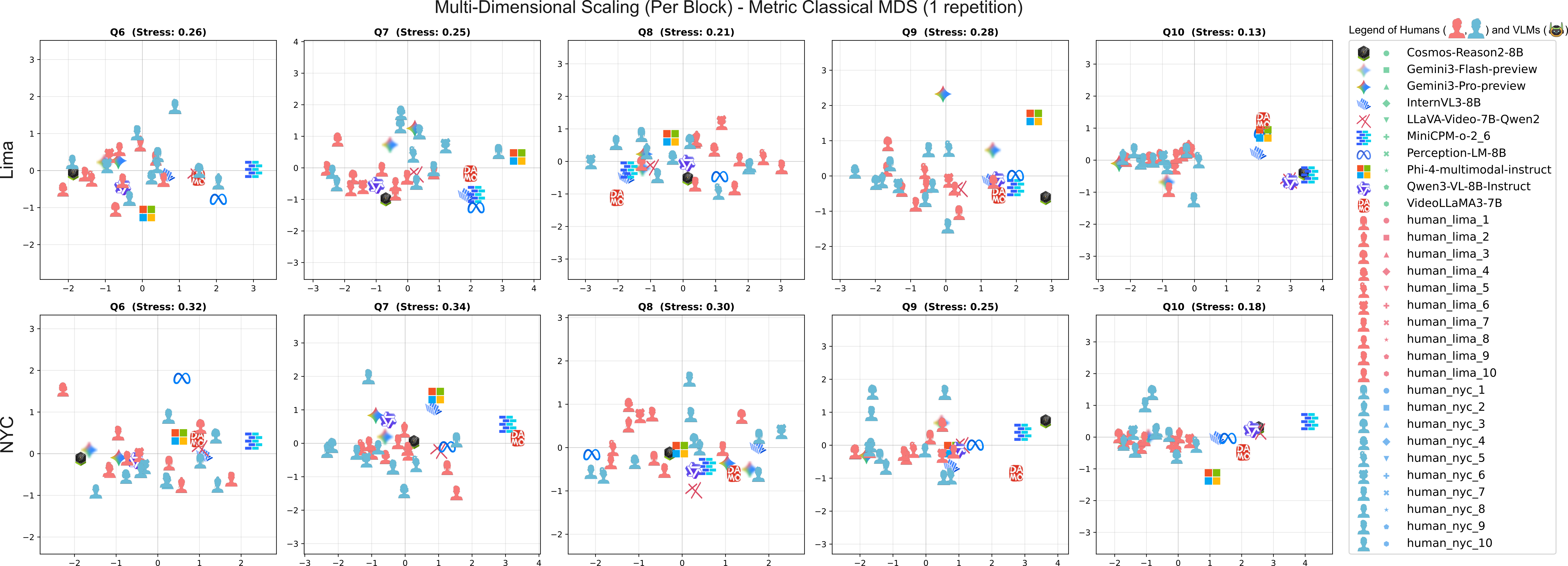}
    \caption{A grid of Multi-Dimensional Scaling plots for Block 2 questions: Q6 to Q10 for both Lima and NYC. Different quality of fits visualized through stress vary per each plot as this is dependent on the dissimilarity score received as input in Figure~\ref{fig:BiasAnalysis} \textit{(B.)}.}
    \label{fig:MDS_Grid}
\end{figure}

\newpage
\clearpage

\begin{figure}[t]
    \centering
    \includegraphics[width=\textwidth]{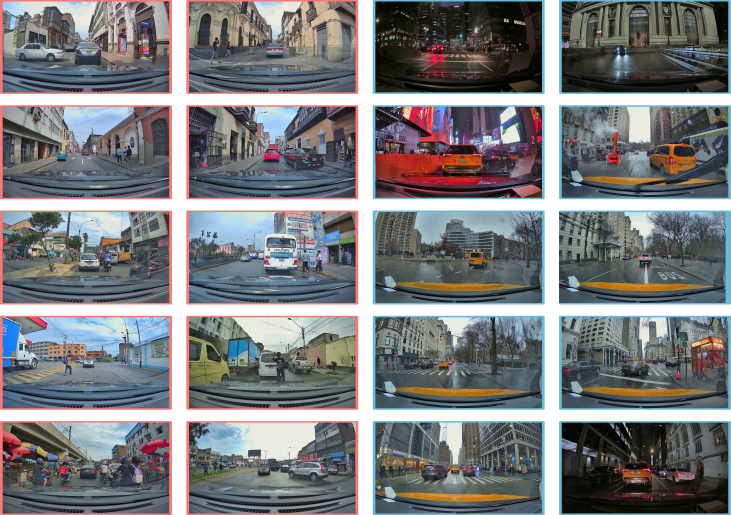}
    \caption{Collage of Stimuli used in the paper (10 scenes are from Lima, and 10 are from NYC). Only 1 frame is shown per video. All videos shown to Humans and VLMs lasted 5 seconds.}
    \label{fig:collage}
\end{figure}

\begin{figure}[t]
    \centering
    \includegraphics[width=\textwidth]{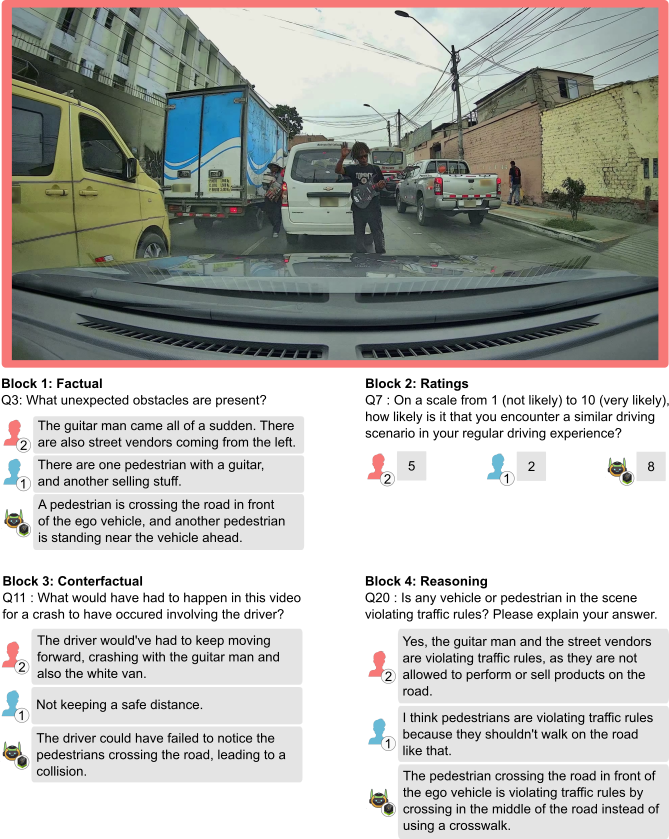}
    \caption{Lima 01 sample \textbf{video}. Notice that \texttt{Cosmos-Reason2-8B} mistakenly thinks that this scene is quite common (Block 2) since it confuses the street vendors for everyday pedestrians (thinking this is a common scene). Further follow-on questions and answers in the sample illustrate how \texttt{Cosmos-Reason2-8B}'s reading of the scene diverges from that of both human groups.}
    \label{fig:showcase_lima_01}
\end{figure}

\begin{figure}[t]
    \centering
    \includegraphics[width=\textwidth]{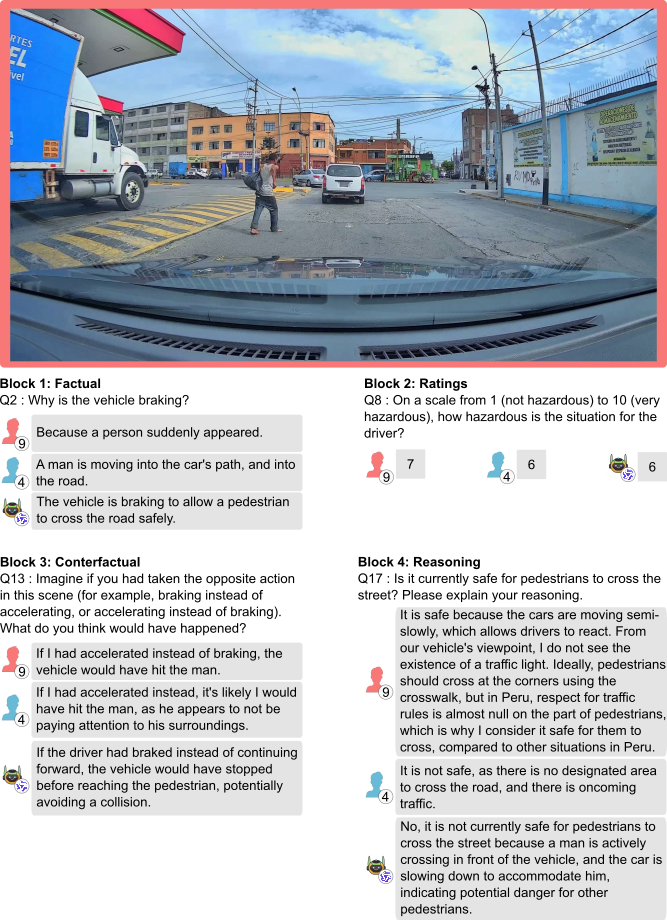}
    \caption{Lima 02 sample \textbf{video}. Here \texttt{Qwen} successfully manages some degree of alignment with the humans from Lima and NYC, although it does not fully seem to understand the subtleness of a pedestrian who is jay walking under the influence.}
    \label{fig:showcase_lima_02}
\end{figure}

\begin{figure}[t]
    \centering
    \includegraphics[width=\textwidth]{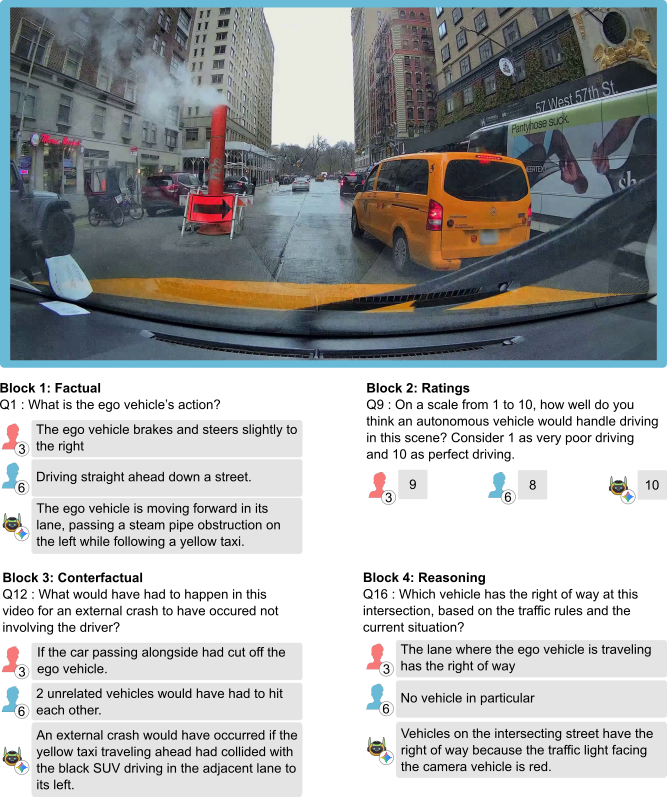}
    \caption{New York 01 sample \textbf{video}. Here \texttt{Google Gemini} shows similar alignment in rating to Humans from Both Lima (red) and New York City (blue). Surprisingly, some of the reasoning answers show levels of confusion as \texttt{Google Gemini} hallucinates a traffic light.}
    \label{fig:showcase_newyork_01}
\end{figure}

\begin{figure}[t]
    \centering
    \includegraphics[width=\textwidth]{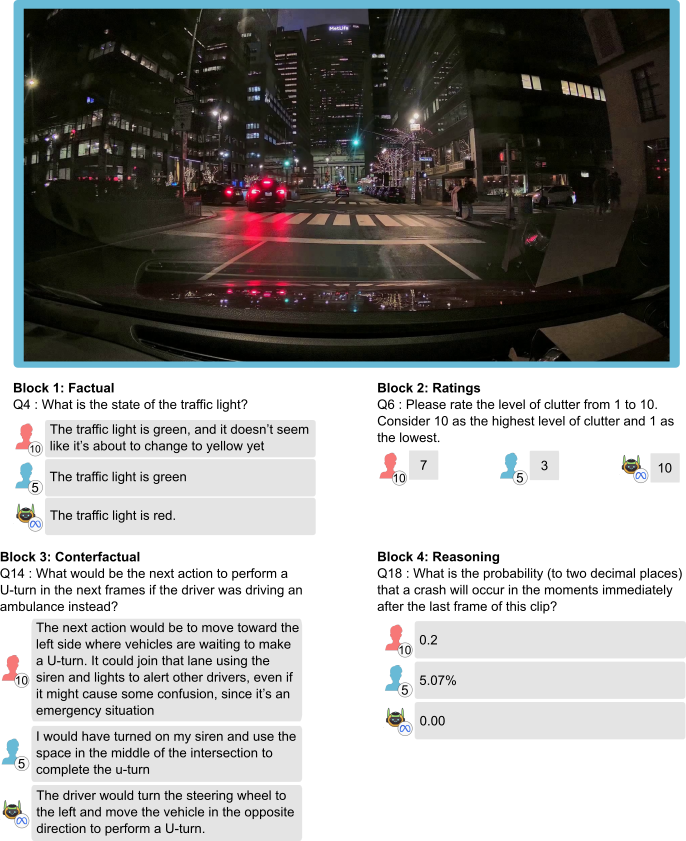}
    \caption{New York 02 sample \textbf{video}. The Factual and Reasoning answers allow us to cognitively probe the \texttt{Meta Perception-LM} model. It disagrees with humans on the color of the traffic light and the probabilistic assignment of a crash to occur.}
    \label{fig:showcase_newyork_02}
\end{figure}

\newpage

\subsection{Inferential Statistical Analysis}
\label{sec:supp-inferential}
The conclusions in the main text---that geography does not modulate the answers, that
Lima and NYC participants respond similarly, and that the Human--VLM gap is large---are
here supported by formal inferential tests rather than visual inspection. All tests are
computed on the already-collected data (no model is re-run). Because geography is a
property of the video, the geography null is tested by permuting the $10/10$ Lima/NYC
video assignment. Table~\ref{tab:supp-factorial} reports the $2\times3$ factorial on the
Block-2 ratings with effect sizes and a video-level power analysis;
Table~\ref{tab:supp-permanova} gives the factorial PERMANOVA on the sentence-embedding
centroids; and Table~\ref{tab:supp-judge} collects the LLM-as-a-Judge robustness checks
(corrected pair counts, family ablation, inference-budget split, and a geography
permutation on the judge scores). The picture is consistent across tests: the
system-type effect is large everywhere, whereas the geography main effect and the
geography$\times$system interaction are negligible in magnitude. The interaction does
reach nominal significance on the Block-2 ratings ($p_{\text{perm}}=0.032$;
PERMANOVA on the Ratings block, $p=0.024$), but it accounts for $0.3\%$ of the
variance (partial $\eta^2=0.003$; $R^2=0.016$) against $10.8\%$ for system type
(partial $\eta^2=0.108$; $R^2=0.586$)---a detectable but practically irrelevant
modulation, and one an $n=10$-per-city design is not powered to characterise further.
The geography main effect is non-significant except on \emph{factual} answers, where it
is expected: factual answers describe scene content, which differs by city. 

\begin{figure}[t]
\centering
\includegraphics[width=\textwidth]{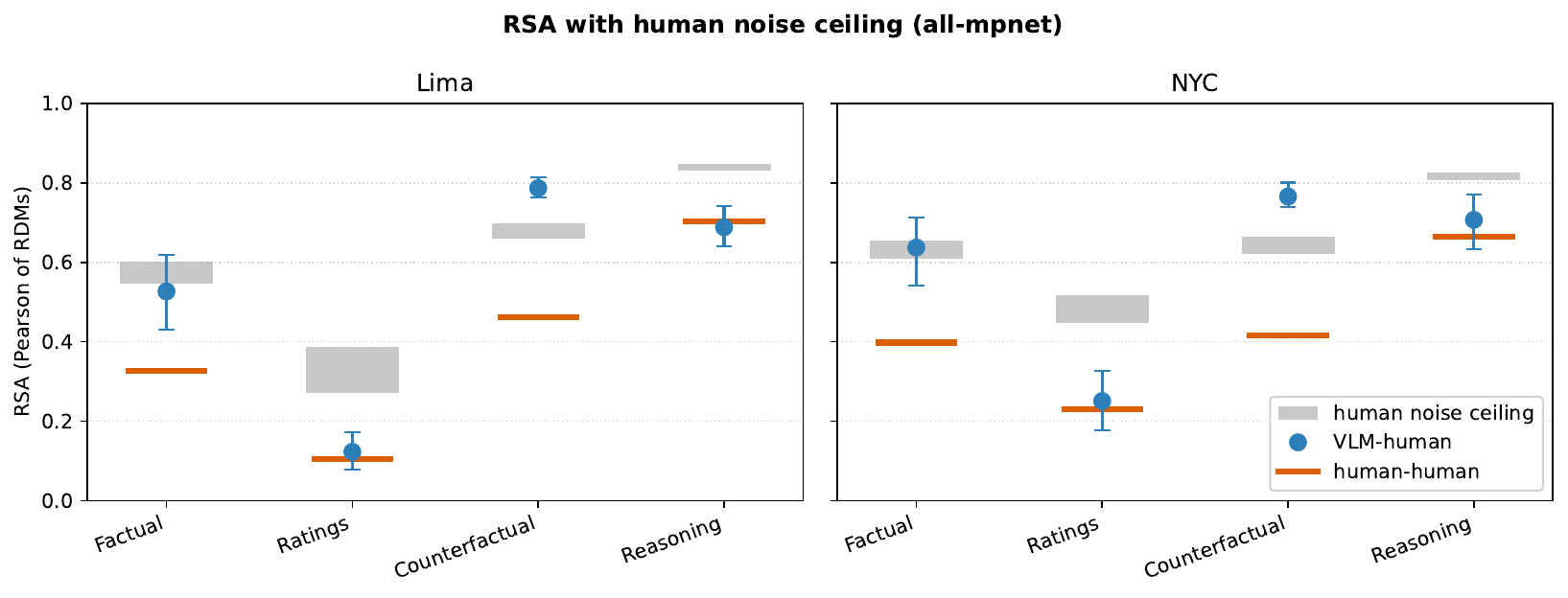}
\caption{RSA with the human noise ceiling (all-mpnet). Grey band = human noise ceiling $[\text{lower},\text{upper}]$ (Nili 2014); blue = VLM$\to$human RSA with bootstrap CI; orange = human--human RSA. The VLM$\to$human alignment falls below the ceiling floor on Ratings and Reasoning, exceeds it on Counterfactual, and lies within the band on Factual.}
\label{fig:supp-noiseceiling}
\end{figure}

Fig.~\ref{fig:supp-noiseceiling} adds the RSA noise ceiling on Ratings and Reasoning the VLM$\to$human alignment sits \emph{below} the human ceiling floor (a genuinely low alignment, not noise), whereas on Counterfactual it \emph{exceeds} the ceiling because humans disagree strongly among themselves; on Factual it lies within the ceiling band.

\begin{table}[t]
\centering
\caption{Inferential test of the $2\times3$ factorial on the Block-2 Likert ratings.
$p$ values are from a label-permutation test that reshuffles the $10/10$ Lima/NYC
assignment of the videos ($10^4$ permutations); geography is a property of the video.
Effect sizes carry video/subject cluster-bootstrap $95\%$ CIs. A video-level power
analysis ($n=10$ per city) had $80\%$ power to detect a Lima--NYC shift of
$0.39$ rating points; the observed shift is $0.078$.}
\label{tab:supp-factorial}
\begin{tabular}{lccc}
\toprule
Term & pseudo-$F$ & $p_{\text{perm}}$ & partial $\eta^2$\\
\midrule
Geography (main)        & 1.00   & 0.55       & 0.0003\\
System type (main)      & 180.2  & $<0.001$   & 0.108\\
Geography $\times$ System & 4.80 & 0.032      & 0.003\\
\midrule
\multicolumn{4}{l}{\textit{Effect sizes (95\% CI)}}\\
Lima--NYC rating gap            & \multicolumn{3}{l}{$-0.078$ pts\ \ $[-0.31,\ 0.16]$}\\
Cohen's $d$: Human$_{\text{Lima}}$ vs VLM        & \multicolumn{3}{l}{$0.75$\ \ $[0.46,\ 1.09]$}\\
Cohen's $d$: Human$_{\text{Lima}}$ vs Human$_{\text{NYC}}$ & \multicolumn{3}{l}{$0.13$\ \ $[-0.05,\ 0.30]$}\\
\bottomrule
\end{tabular}
\end{table}

\begin{table}[t]
\centering
\caption{Factorial PERMANOVA on the sentence-embedding centroids
(all-mpnet, cosine distance; $R^2$ = fraction of variance explained, $2000$
permutations). System type dominates every block while the
geography$\times$system interaction---the actual cross-cultural quantity---is
$\approx 0$ throughout. The geography main effect appears only for \emph{factual}
answers, as expected: factual answers describe scene content, which differs by city.}
\label{tab:supp-permanova}
\begin{tabular}{lcccccc}
\toprule
 & \multicolumn{2}{c}{Geography} & \multicolumn{2}{c}{System type} & \multicolumn{2}{c}{Geo.$\times$System}\\
\cmidrule(lr){2-3}\cmidrule(lr){4-5}\cmidrule(lr){6-7}
Block & $R^2$ & $p$ & $R^2$ & $p$ & $R^2$ & $p$\\
\midrule
Factual        & 0.205 & $<0.001$ & 0.272 & $<0.001$ & 0.000 & 1.00\\
Ratings        & 0.004 & 0.18     & 0.586 & $<0.001$ & 0.016 & 0.024\\
Counterfactual & 0.019 & $<0.001$ & 0.249 & 0.001    & 0.001 & 0.39\\
Reasoning      & 0.034 & $<0.001$ & 0.613 & $<0.001$ & 0.000 & 1.00\\
\bottomrule
\end{tabular}
\end{table}

\begin{table}[t]
\centering
\caption{Robustness of the LLM-as-a-Judge analysis. \textit{(a)} The deduplicated
pipeline produces $43{,}500=\binom{30}{2}\times100$ pairs per block; the
comparability (Stage-1) rate is $79$--$85\%$. The larger counts in the original
text correspond to an older ordered Cartesian run. \textit{(b)} The judge scores
models from its own (Qwen) family more generously; we report agreement with and
without that family. \textit{(c)} Splitting by inference budget reveals strong
heterogeneity. \textit{(d)} Geography is non-significant for the judge scores in
every block (video-level permutation).}
\label{tab:supp-judge}
\begin{tabular}{lccc}
\toprule
\multicolumn{4}{l}{\textit{(a) Pairs per block}\quad (comparable / total, rate)}\\
\midrule
Block & comparable & total & vs.\ reported\\
Factual        & 36{,}943 (85.0\%) & 43{,}500 & 63{,}121\\
Counterfactual & 34{,}411 (79.1\%) & 43{,}500 & 57{,}886\\
Reasoning      & 36{,}396 (83.7\%) & 43{,}500 & 61{,}302\\
\midrule
\multicolumn{4}{l}{\textit{(b) Judge family ablation}\quad (mean agreement $J$)}\\
\midrule
Human--VLM, with Qwen family    & \multicolumn{3}{l}{$0.218$}\\
Human--VLM, without Qwen family & \multicolumn{3}{l}{$-0.004$}\\
Overall $J$ (all / excl.\ Qwen family) & \multicolumn{3}{l}{$0.161$ / $0.142$}\\
\midrule
\multicolumn{4}{l}{\textit{(c) Inference-budget split}\quad (agreement $J$)}\\
\midrule
Gemini (1\,fps, $T{=}1.0$): within / vs.\ human & \multicolumn{3}{l}{$0.90$ / $0.36$}\\
Open (10\,fps, $T{=}0.5$): within / vs.\ human  & \multicolumn{3}{l}{$0.34$ / $-0.04$}\\
\midrule
\multicolumn{4}{l}{\textit{(d) Geography permutation on judge scores}}\\
\midrule
Factual / Counterfactual / Reasoning & \multicolumn{3}{l}{$p=0.32$ / $0.12$ / $0.22$}\\
\bottomrule
\end{tabular}
\end{table}

\begin{figure}[t]
\centering
\includegraphics[width=\textwidth]{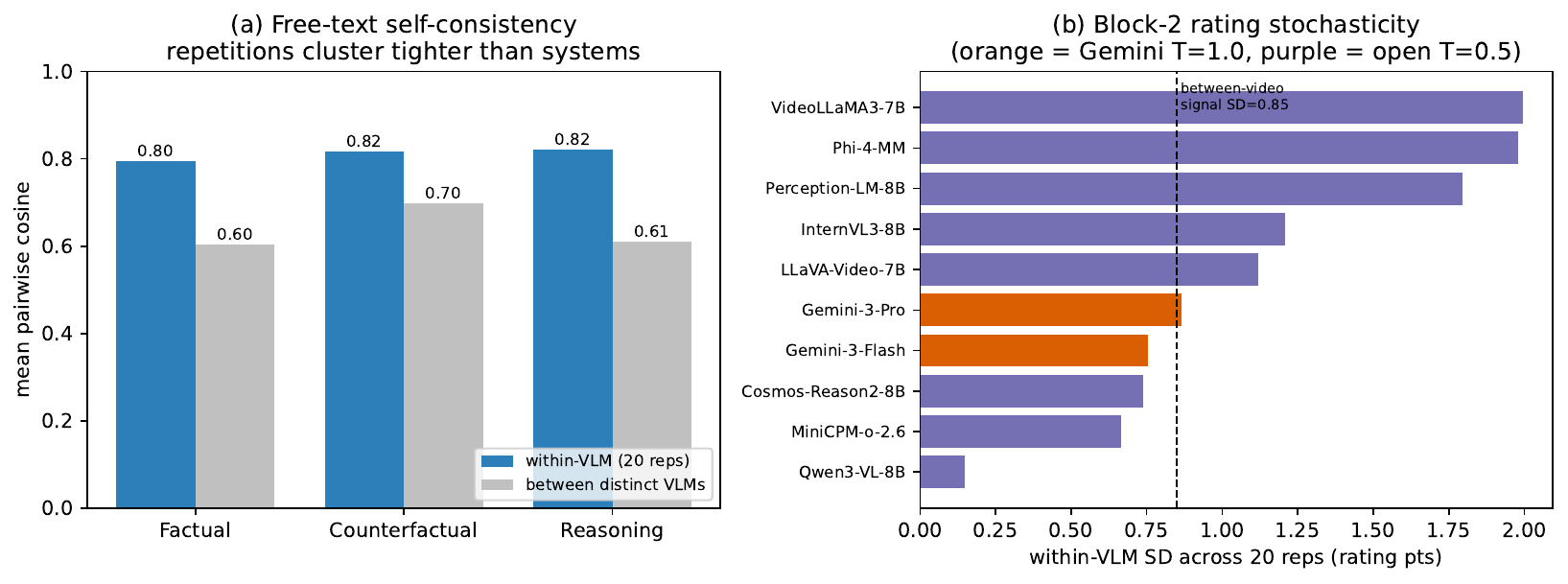}
\caption{Repetition analysis. \textit{(a)} Free-text semantic self-consistency
(within-VLM, over $20$ repetitions) vs.\ between-VLM similarity, by block. \textit{(b)}
Per-VLM within-repetition standard deviation of the Block-2 ratings; orange = Gemini
($T{=}1.0$), purple = open models ($T{=}0.5$); dashed line = between-video signal SD.}
\label{fig:supp-repcons}
\end{figure}

Each VLM answered every (video, question) pair $20$ times whereas each participant
answered once; all analyses in the paper use a single VLM response (the first
repetition), matching the single human response. Table~\ref{tab:supp-reprobust} shows the
Block-2 factorial is unchanged whether we use the first repetition or the mean of the
$20$ repetitions, and Table~\ref{tab:supp-repcons} (with Fig.~\ref{fig:supp-repcons})
characterizes the discarded repetitions: within-VLM repetitions cluster far tighter than
distinct systems, and this run-to-run variability is itself geography-invariant.

\begin{table}[t]
\centering
\caption{Robustness of the Block-2 factorial to the single-repetition choice: results
are essentially identical using the first repetition (as in the paper) or the mean of
each VLM's $20$ repetitions. If anything, averaging sharpens the system effect and
further weakens geography.}
\label{tab:supp-reprobust}
\begin{tabular}{lcc}
\toprule
Block-2 factorial term & rep.\ 1 (paper) & mean of 20 reps\\
\midrule
Geography $p_{\text{perm}}$ & 0.554 & 0.708\\
Geography partial $\eta^2$ & 0.0003 & 0.0001\\
System $p_{\text{perm}}$ & $<0.001$ & $<0.001$\\
System partial $\eta^2$ & 0.108 & 0.125\\
Geo.$\times$System $p_{\text{perm}}$ & 0.032 & 0.028\\
Lima--NYC gap (pts) & -0.078 & -0.044\\
Cohen's $d$ Human--VLM & 0.752 & 0.837\\
Cohen's $d$ Lima--NYC humans & 0.132 & 0.132\\
\bottomrule
\end{tabular}
\end{table}

\begin{table}[t]
\centering
\caption{Within-VLM repetition consistency on the already-collected data.
\textit{(a)} In the free-text blocks a VLM's $20$ repetitions are far more similar to one
another than distinct VLMs are, and this is geography-invariant. \textit{(b)} Numeric
Block-2 ratings are stochastic across repetitions, which is why Block-2 conclusions are
drawn from full distributions; this stochasticity is also geography-invariant.}
\label{tab:supp-repcons}
\begin{tabular}{lccc}
\toprule
\multicolumn{4}{l}{\textbf{(a) Free-text semantic self-consistency (mean pairwise cosine)}}\\
\midrule
Block & within-VLM (20 reps) & between distinct VLMs & --\\
\midrule
Factual & 0.795 & 0.604 & \\
Counterfactual & 0.817 & 0.699 & \\
Reasoning & 0.821 & 0.611 & \\
\midrule
All free-text & 0.811 & 0.638 & \\
Lima vs NYC & 0.810 / 0.812 & \multicolumn{2}{l}{perm.\ $p=0.70$ (no geo.\ effect)}\\
\midrule
\multicolumn{4}{l}{\textbf{(b) Block-2 numeric run-to-run stochasticity (rating points, 1--10)}}\\
\midrule
Mean within-VLM rep.\ SD & \multicolumn{3}{l}{1.13}\\
Between-video signal SD & \multicolumn{3}{l}{0.85}\\
Gemini (T=1.0) vs open (T=0.5) & \multicolumn{3}{l}{0.81 vs 1.21}\\
Lima vs NYC rep.\ SD & \multicolumn{3}{l}{1.13 / 1.12, perm.\ $p=0.76$}\\
\bottomrule
\end{tabular}
\end{table}

\paragraph{Response length and judge rubric.} Response register differs by group (Kruskal--Wallis per block, all $p<10^{-40}$; e.g.\ Reasoning mean word counts $25.2$ for VLMs vs $12.7$/$7.4$ for Lima/NYC humans), yet this does not drive the judge: the absolute word-count gap between two answers barely predicts their agreement score (Spearman $\rho=0.034$ overall, $-0.011$ on Human--VLM pairs) and only weakly the comparability decision ($\rho=-0.088$), so the Human--VLM divergence is not a length artefact. The judge rubric is likewise robust: recoding the rare $-1$ (partial contradiction) to neutral shifts overall agreement only from $J=0.161$ to $0.201$ and per-block means by $<0.04$, so the conclusions are stable to the scoring scheme.

\end{document}